\documentclass[journal]{IEEEtran}

\usepackage{graphicx} 
\usepackage{subfigure} 

\usepackage[nocompress]{cite}

\usepackage{algorithm}
\usepackage{algorithmic}
\usepackage{amsmath}
\usepackage{amssymb}
\usepackage{bm}

\usepackage{hyperref}

\usepackage{diagbox}
\usepackage{arydshln}
\usepackage{listings}
\usepackage{multirow}

\newcommand{\experiment}{ZSL}
\newcommand{\deepmodel}{SR2CNN}

\newtheorem{remark}{Remark} 

\begin{document} 
\title{SR2CNN: Zero-Shot Learning for Signal Recognition}

\author{Yihong Dong,
        Xiaohan Jiang,
        Huaji Zhou,
        Yun Lin,
        and~Qingjiang Shi
\thanks{This work was supported in part by the National Key Research and Development Project under grant 2017YFE0119300, and in part by the NSFC under Grants 61731018 and U1709219. \emph{(Corresponding author: Qingjiang Shi)}}
\thanks{Y. Dong, X. Jiang and Q. Shi are all with the School of
Software Engineering, Tongji University, Shanghai 201804, China. Q. Shi is also with the Shenzhen Research Institute of Big Data, Shenzhen 518172, China. (e-mail: 1931543@tongji.edu.cn; 1752916@tongji.edu.cn; shiqj@tongji.edu.cn)}
\thanks{H. Zhou is with the School of Artificial Intelligence, Xidian University, Xi'an 710071, China (e-mail:  zhouhuaji1988@sina.com)}
\thanks{Y. Lin is with the School of College of Information and Communication Engineering, Harbin Engineering University, Harbin 150001, China (e-mail:  linyun@hrbeu.edu.cn)}
}


\maketitle

\begin{abstract} 
Signal recognition is one of the significant and challenging tasks in the signal processing and communications field. It is often a common situation that there's no training data accessible for some signal classes to perform a recognition task. Hence, as widely-used in image processing field, zero-shot learning (ZSL) is also very important for signal recognition. Unfortunately, ZSL regarding this field has hardly been studied due to inexplicable signal semantics. This paper proposes a ZSL framework, signal recognition and reconstruction convolutional neural networks (SR2CNN), to address relevant problems in this situation. The key idea behind SR2CNN is to learn the representation of signal semantic feature space by introducing a proper combination of cross entropy loss, center loss and reconstruction loss, as well as adopting a suitable distance metric space such that semantic features have greater minimal inter-class distance than maximal intra-class distance. The proposed SR2CNN can discriminate signals even if no training data is available for some signal class. Moreover, SR2CNN can gradually improve itself in the aid of signal detection, because of constantly refined class center vectors in semantic feature space. These merits are all verified by extensive experiments with ablation studies.
\end{abstract} 


\IEEEpeerreviewmaketitle

\section{Introduction}
\IEEEPARstart{N}{owadays}, developments in deep convolutional neural networks (CNNs) have made remarkable achievement in the area of signal recognition, improving the state of the art significantly, such as \cite{o2018over,gama2018convolutional,peng2018modulation,o2016unsupervised,du2012noise} and so on. Generally, a vast majority of existing learning methods follow a closed-set assumption\cite{garriga2008closed}, that is, all of the test classes are assumed to be the same as the training classes. However, in the real-world applications, new signal categories often appear while the model is only trained for the current dataset with some limited known classes. 
It is open-set learning \cite{scheirer2012toward,bendale2016towards,liu2019separate,geng2020collective} that was proposed to partially tackle this issue (i.e., test samples could be from unknown classes). The goal of an open-set recognition system is to reject test samples from unknown classes while maintaining
the performance on known classes. However, in some cases, the learned model should be able to not only differentiate the unknown classes from known classes, but also distinguish among different unknown classes.  Zero-shot learning (ZSL) \cite{palatucci2009zero,socher2013zero,wang2020network} is one way to address the above challenges and has been applied in image tasks. For images, it is easy for us to extract some human-specified high-level descriptions as semantic attributes. For example, from a picture of zebra, we can extract the following semantic attributes 1) color: white and black, 2) stripes: yes, 3) size: medium, 4) shape: horse, 5) land: yes. However, for a real-world signal it is almost impossible to have a high-level description due to obscure signal semantics. Therefore, \emph{although ZSL has been widely used in image tasks, to the best of our knowledge it has not yet been studied for signal recognition.}\footnote{A closely related work is \cite{gao2019zero} which proposed a ZSL method for fault diagnosis based on vibration signal. Notice that fault diagnosis is a binary classification problem, which is different from the multi-class signal recognition. More importantly, the ZSL definition in this paper is standard and quite different from the ZSL definition of \cite{gao2019zero}, where ZSL refers to fault diagnosis with unknown motor loads and speeds, which is essentially domain adaptation, while in our paper, ZSL refers to recognition of unknown classes of the signal.} 

\begin{figure}[htbp]
\begin{center}
\centerline{\includegraphics[width=9cm]{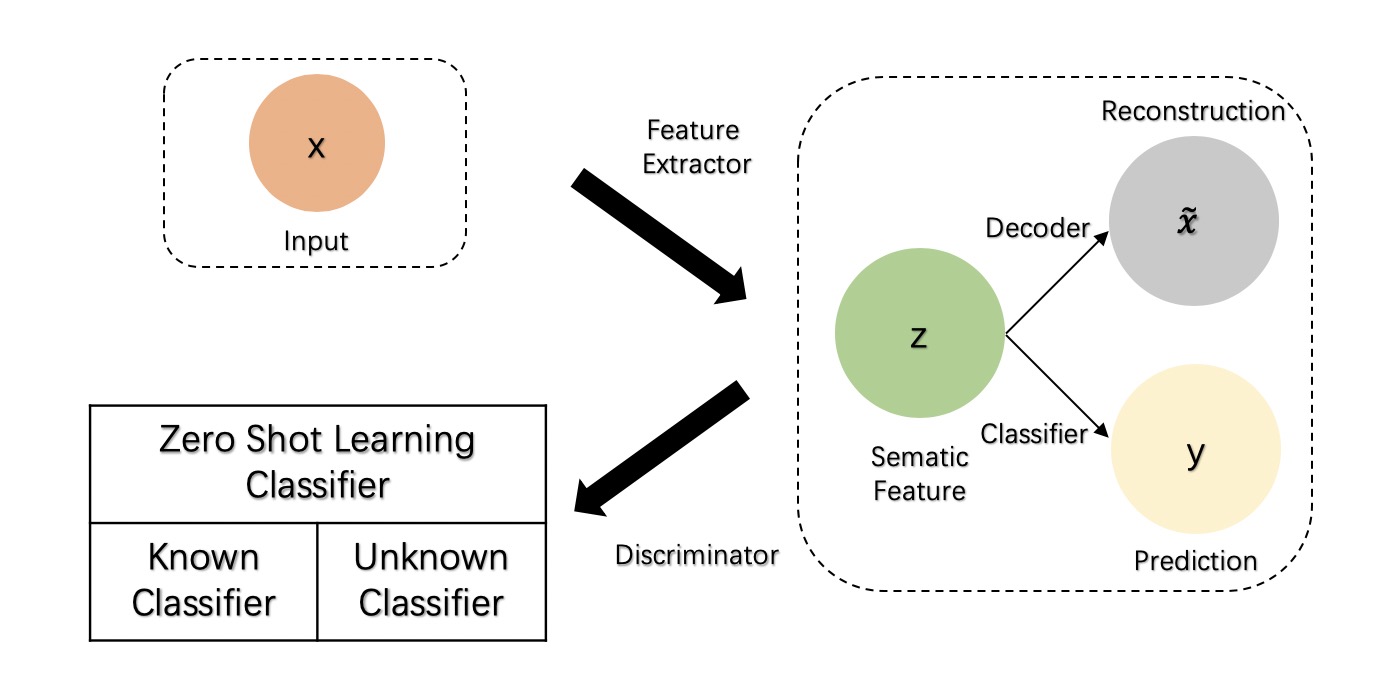}}
\caption{Overview of \deepmodel. In \deepmodel, a pre-processing (top left) transforms signal data to input $x$. A deep net (right) is trained to provide semantic feature $z$ within known classes while maintaining the performance on decoder and classifier according to reconstruction $\tilde{x}$ and prediction $y$. A zero-shot learning classifier, which consists of a known classifier and an unknown classifier, exploits $z$ for discriminator.}
\label{1}
\end{center}
\end{figure}

\begin{figure*}[htbp]
\begin{center}
\centerline{\includegraphics[width=17cm]{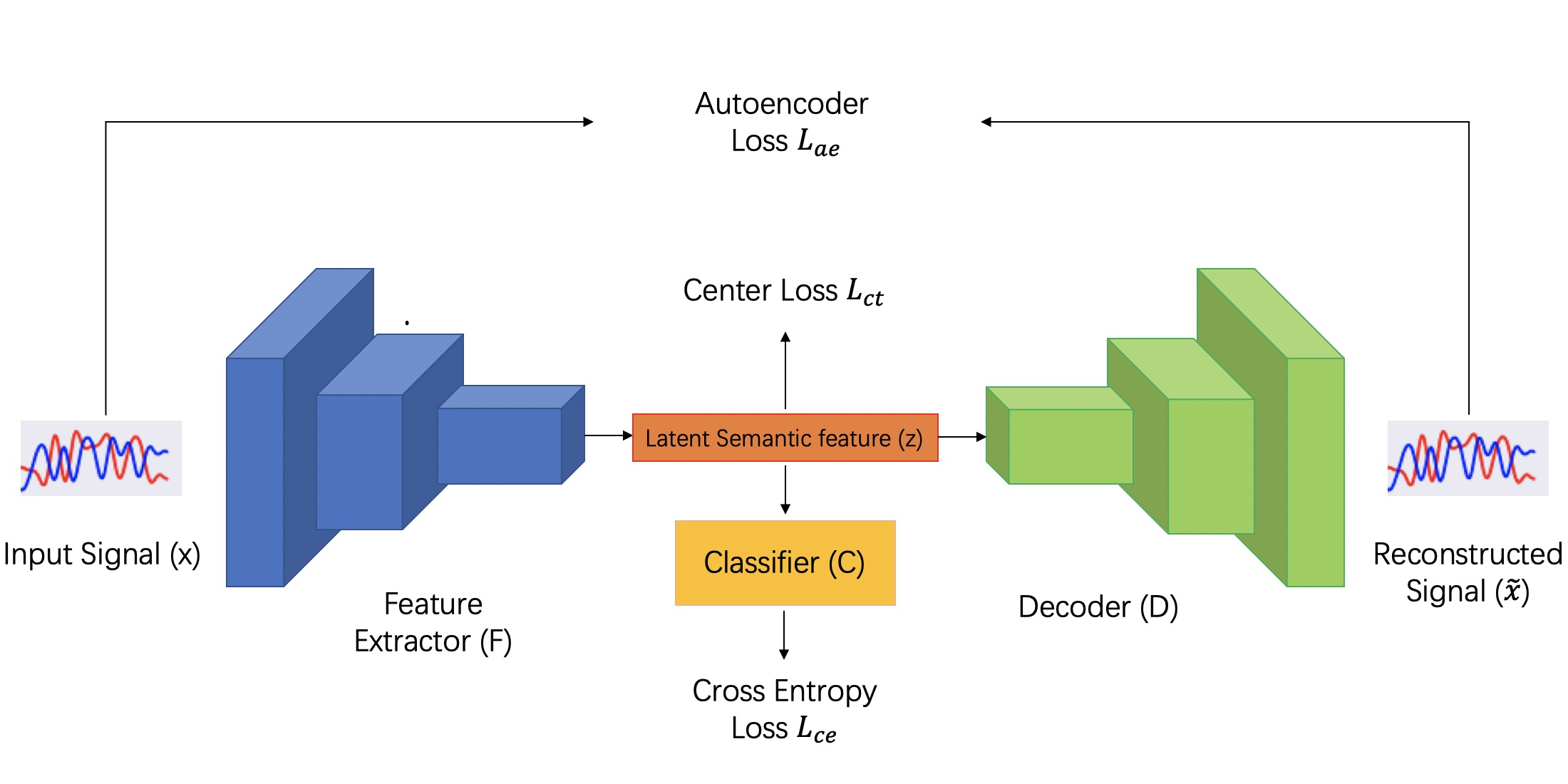}}
\caption{The architecture of feature extractor ($F$), classifier ($C$) and decoder ($D$). $F$ takes any input signal $x$ and produces a latent semantic feature $z$. $z$ is used by $C$ and $D$ to predict class label and to reconstruct the signal $\widetilde{x}$, respectively. The $L_{ce}$, $L_{ct}$ and $L_{r}$ are calculated on training these networks.}
\label{2}
\end{center}
\end{figure*}

\emph{In this paper, unlike the conventional signal recognition task where a classifier is learned to distinguish only known classes (i.e., the labels of test data and training data are all within the same set of classes), we aim to propose a learning framework that can not only classify known classes but also unknown classes without annotations.} To do so, a key issue that needs to be addressed is to automatically learn a representation of semantic attribute space of signals. In our scheme, CNN combined with autoencoder is exploited to extract the semantic attribute features. Afterwards, semantic attribute features are well-classified using a suitably defined distance metric. 
The overview of proposed scheme is illustrated in Fig. \ref{1}. 

In addition, to make a self-evolution learning model, incremental learning needs to be considered when the algorithm is executed continuously. The goal of incremental learning is to dynamically adapt the model to new knowledge from newly coming data without forgetting the already learned one. Based on incremental learning,  the obtained model will gradually improve its performance over time. 

In summary, the main contribution of this paper is threefold:
\begin{itemize}
    \item First, we propose a deep CNN-based zero-shot learning framework, called \deepmodel, for open-set signal recognition. \deepmodel~is trained to extract semantic feature $z$ while maintaining the performance on decoder and classifier. Afterwards, the semantic feature $z$ is exploited to discriminate signal classes.
    \item Second, extensive experiments on various signal datasets show that the proposed \deepmodel~can discriminate not only known classes but also unknown classes and it can gradually improve itself.
    \item Last but not least, we provide a new signal dataset SIGNAL-202002 including eight digital and three analog modulation classes.
\end{itemize}


\section{Related Work}
In recent years, signal recognition via deep learning has achieved a series of successes. O’Shea et al. \cite{o2016convolutional} proposed the convolutional radio modulation recognition networks, which can adapt itself to the complex temporal radio signal domain, and also works well at low signal-to-noise ratios (SNRs). 
The work \cite{o2018over} used residual neural network \cite{he2016deep} to perform the signal recognition tasks across a range of configurations and channel impairments, offering referable statistics. Peng et al. \cite{peng2018modulation} used two convolutional neural networks, AlexNet and GoogLeNet, to address modulation classification tasks, demonstrating the significant advantage of deep learning based approach in this field. The authors in \cite{zheng2018big} presented a deep learning based big data processing architecture for end-to-end signal processing task, seeking to obtain important information from radio signals. The work presented in \cite{flowers2019evaluating} evaluated the adversarial evasion attacks that causes the misclassification in the context of wireless communications. In \cite{duan2018automatic}, the authors proposed an automatic multiple multicarrier waveforms classification and used the principal component analysis to suppress the additive white Gaussian noise and reduce the input dimensions of CNNs. Additionally, the work \cite{wong2019specific} proposed a specific emitter identification using CNN-Based inphase/quadrature (I/Q) imbalance estimators. The work \cite{huang2019automatic} proposed a compressive convolutional neural network for automatic modulation classification. In \cite{dorner2017deep}, the authors used unsynchronized off-the-shelf software-defined radios to build up a complete communications system which is solely composed of deep neural networks, demonstrating that over-the-air transmissions are possible. 

Moreover, the work \cite{hoang2019automatic} proposed an LPI radar waveform recognition technique based on a single-shot multi-box detector and a supplementary classifier.
The work \cite{vanhoy2018hierarchical} proposed a more flexible network architecture with an augmented hierarchical-leveled training techniques to decently classify a total of 29 signals. 
O’Shea et al. \cite{o2018demonstrating} used both the auto-encoder-based communications system and the feature learning-based radio signal sensor to emulate the optimization procedure directly on real-world data samples and distributions.
Baldini et al. \cite{baldini2018comparison} utilized various techniques to transform the time series derived from the radio frequency to images, then applied a deep CNN to conduct the identification task, finally outperforming those conventional dissimilarity-based methods.
The work \cite{hiremath2019deep} trained a convolutional neural network on time and stockwell channeled images for radio modulation classification tasks, performing superior to those networks trained on just raw I-Q time series samples or time-frequency images.
The authors for \cite{zhang2019deep} demonstrated the generality of the effectiveness of deep learning at the interference source identification task, while using band selection, SNR selection and sample selection to optimize training time.
The work \cite{sankhe2019no} presented a novel system based on CNNs to “fingerprint” a unique radio from a large pool of devices by deep-learning the fine-grained hardware impairments imposed by radio circuitry on physical-layer I/Q samples.
The work \cite{liang2019towards} proposed a DNN based power control method that aims at solving the non-convex optimization problem of maximizing the sum rate of a fading multi-user interference channel.
Chen et al. \cite{chen2020data} proposed adaptive transmission scheme and generalized data representation scheme to address the limited data rate issue.
In \cite{roy2019rfal}, the authors proposed the radio frequency (RF) adversarial learning framework for building a robust system to identify rogue RF transmitters by designing and implementing a generative adversarial net. 
The work \cite{tan2020intelligent} presented an intelligent duty-cycle medium access control protocol to realize the effective and fair spectrum sharing between LTE and WiFi systems without requiring signalling exchanges.

For semi-supervised learning, the work \cite{li2018generative} proposed a generative adversarial networks-based automatic modulation recognition for cognitive radio networks. Besides, when it comes to unsupervised learning, the authors in \cite{usama2019unsupervised} provided a comprehensive survey of the applications of unsupervised learning in the domain of networking, offering certain instructions. The work \cite{tang2018implementation} built an automatic modulation recognition architecture, based on stack convolution autoencoder, using the reconfigurability of field-programmable gate arrays. These experiments basically follow closed-set assumption, namely, their deep models are expected to, whilst are only capable to distinguish among already-known signal classes.

All the above works cannot handle the case with unknown signal classes. When considering the recognition task of those unknown signal classes, some traditional machine learning methods like anomaly (also called outlier or novelty) detection can more or less provide some guidance. Isolation Forest \cite{liu2008isolation} constructs a binary search tree to preferentially isolate those anomalies. Elliptic Envelope \cite{rousseeuw1999fast}, fits an ellipse for enveloping these central data points, while rejecting the outsiders. One-class SVM \cite{chen2001one}, an extension of SVM, finds a decision hyperplane to separate the positive samples and the outliers. Local Outlier Factor \cite{breunig2000lof}, uses distance and density to determine whether a data point is abnormal or not. The work \cite{yoshihashi2019classification} proposed a classification-reconstruction learning for open-set recognition method that utilizes latent representations for reconstruction and enables robust unknown detection without harming the known-class classification accuracy. Geng et al. \cite{geng2020recent} provided a comprehensive survey of existing open set recognition techniques covering various aspects ranging from related definitions, representations of models, datasets, evaluation criteria, and algorithm comparisons. The work \cite{liu2020few} proposed a multitask deep learning method that simultaneously conducts classification and reconstruction in the open world where unknown classes may exist. Moreover, the work \cite{roy2019detection} proposed a generative adversarial networks based technique to address an open-set problem, which is to identify rogue RF transmitters and classify trusted ones. The work \cite{rajendran2018saife} presented spectrum anomaly detector with interpretable features, which is an adversarial autoencoder based unsupervised model for wireless spectrum anomaly detection. The above open-set learning methods can indeed identify known samples (positive samples) and detect unknown ones (outliers). However, a common and inevitable defect of these methods are that they can never carry out any further classification tasks for the unknown signal classes. 

Zero-shot learning is well-known to be able to classify unknown classes and it has already been widely used in image tasks. For example, the work \cite{palatucci2009zero} proposed a ZSL framework that can predict unknown classes omitted from a training set by leveraging a semantic knowledge base. Another paper \cite{socher2013zero} proposed a novel model for jointly doing standard and ZSL classification based on deeply learned word and image representations. The efficiency of ZSL in image processing field majorly profits from the perspicuous semantic attributes which can be manually defined by high-level descriptions. However, it is almost impossible to give any high-level descriptions regarding signals and thus the corresponding semantic attributes cannot be easily acquired beforehand. This may be the main reason why ZSL has not yet been studied in signal recognition. 

\begin{figure*}[htbp]
\begin{center}
\centering 
  \subfigure[Max unpooling]{ \label{3:a}
    \includegraphics[width=5.5cm]{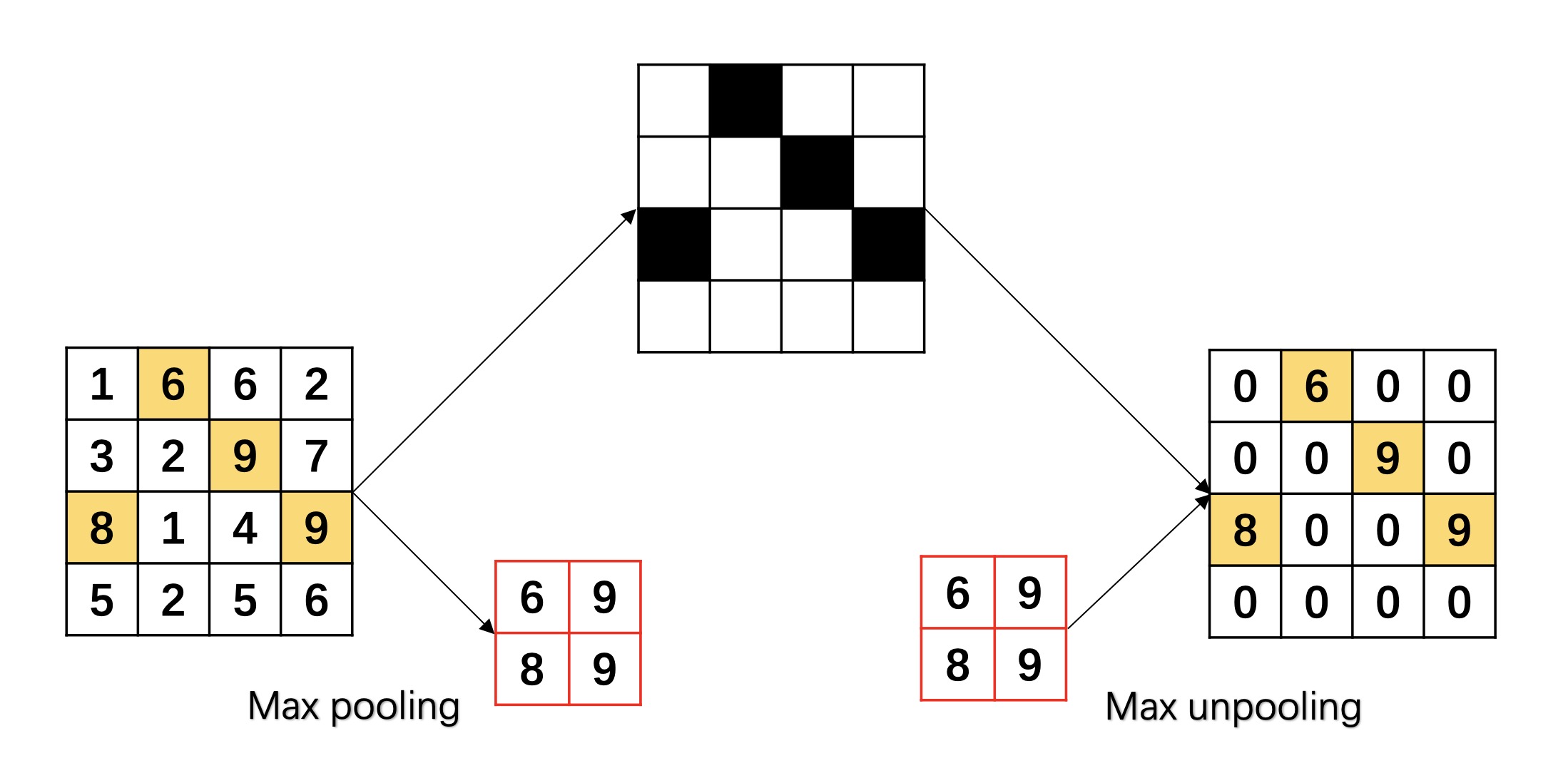} 
  } 
  \subfigure[Average unpooling]{ \label{3:b}
    \includegraphics[width=5.5cm]{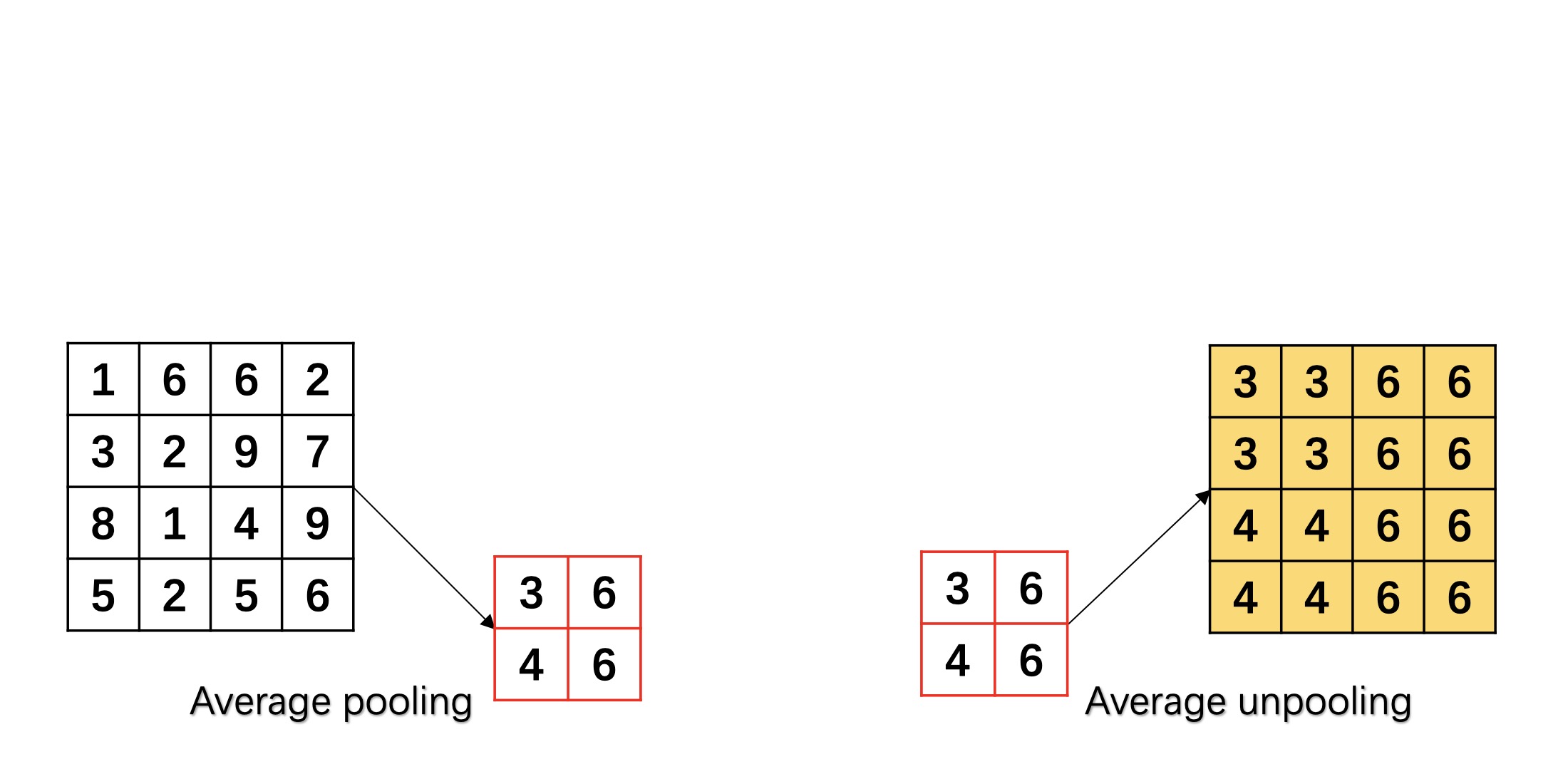} 
  }
  \subfigure[Deconvolution]{ \label{3:c}
    \includegraphics[width=5.5cm]{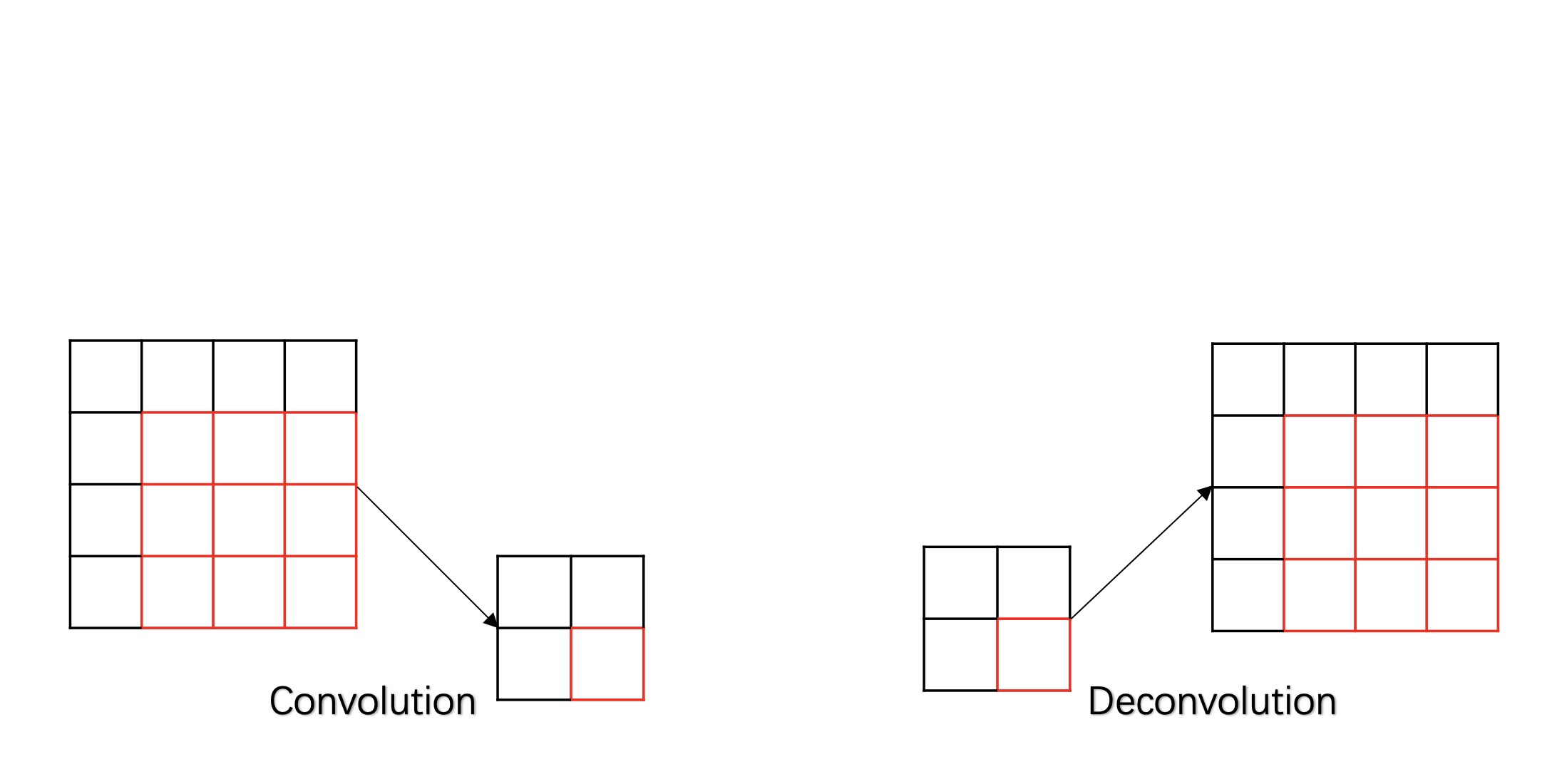} 
  }
\caption{The diagrams of max unpooling, average unpooling and deconvolution. (a) Max unpooling with grid of $2\times2$, where the stride and padding are 2 and 0. (b) Average unpooling with grid of $2\times2$, where the stride and padding are 2 and 0. (c) Deconvolution with kernel of $3\times3$, where the stride and padding are 1 and 0 respectively.}
\end{center}
\end{figure*}

\section{Problem Definition}
We begin by formalizing the problem. Let $X$, $Y$ be the signal input space and output space. The set $Y$ is partitioned into $K$ and $U$, denoting the collection of known class labels and unknown labels, respectively. 

Given training data $\{(x_1,y_1),\hdots,(x_n,y_n)\} \subset X \times K$, the task is to extrapolate and recognize signal class that belongs to $Y$. Specifically, when we obtain the signal input data $x \in X$, the proposed learning framework, elaborated in the sequel, can rightly predict the label $y$. Notice that our learning framework differs from open-set learning in that we not only classify the $x$ into either $K$ or $U$, but also predict the label $y \in Y$. Note that $Y$ includes both known classes $K$ and unknown classes $U$. 

We restrict our attention to ZSL that uses semantic knowledge to recognize $K$ and extrapolate to $U$. To this end, we first map from $X$ into the semantic space $Z$, and then map this semantic encoding to a class label. Mathematically, we can use nonlinear mapping to describe our scheme as follows. $H$ is the composition of two other functions, $F$ and $P$ defined below, such that: 
\begin{equation}
\begin{aligned}
&H = P(F(\cdot)) \\
&F : X \rightarrow Z \\
&P : Z \rightarrow Y
\end{aligned}
\end{equation}
Therefore, our task is left to find proper $F$ and $P$ to build up a learning framework that can identify both known signal classes and unknown signal classes.

\section{Proposed Approach}
\label{Approach}
This section formally presents a non-annotation zero-shot learning framework for signal recognition. Overall, the proposed framework is mainly composed of the following four modules:
\begin{enumerate}
    \item Feature Extractor ($F$)
    \item Classifier ($C$)
    \item Decoder ($D$)
    \item Discriminator ($P$)
\end{enumerate}

Our approach consists of two main steps. In the first step, we build a semantic space for signals through $F$, $C$ and $D$. Fig. \ref{2} shows the architecture of $F$, $C$ and $D$. $F$ is modeled by a CNN architecture that projects the input signal onto a latent semantic space representation. $C$, modeled by a fully-connected neural network, takes the latent semantic space representation as input and determines the label of data. $D$, modeled by another CNN architecture, aims to produce the reconstructed signal which is expected to be as similar as possible to the input signal. In the second step, we find a proper distance metric for the trained semantic space and use the distance to discriminate the signal classes. $P$ is devised to discriminate among all classes including both known and unknown.

\subsection{Feature Extractor, Classifier and Decoder}
Signal is a special data type, which is very different from image. While it is easy to give a description of semantic attributes of images in terms of visual information, extracting semantic features of signals without relying on any computation is almost impossible. Hence, a natural way to automatically extract the semantic information of signal data is using feature extractor networks $F$. Considering about the unique features of signals, the input shape of $F$ should be a rectangle matrix with 2 rows rather than square matrix. In our scheme, $F$ consists of four convolutional layers and two fully connected layers.

Generally, $F$ can be represented by a mapping from the input space $X$ to the latent semantic space $Z$. In order to minimize the intra-class variations in space $Z$ while keeping the inter-classes' semantic features well separated, center loss \cite{wen2016discriminative} is used. Let $x_i \in X$ and $y_i$ be the label of $x_i$, then $z_i = F(x_i) \in Z$. Assuming that batch size is $N$, the center loss is expressed as follows:
\begin{equation} \label{ct}
    L_{ct} = \frac{1}{2N}\sum_{i=1}^N ||F(x_i) - c_{y_i}||^2_2
\end{equation}
where $c_{y_i}$ denotes the semantic center vector of class $y_i$ in $Z$ and the $c_{y_i}$ needs to be updated as the semantic features of class $y_i$ changed. Ideally, entire training dataset should be taken into account and the features of each class need to be averaged in every iterations. In practice, $c_{y_i}$ can be updated for each batch according to  $c_{y_i}\leftarrow  c_{y_i}-\alpha \Delta_{c_{y_i}}$, where $\alpha$ is the learning rate and $\Delta_{c_{y_i}}$ is computed via
\begin{equation}
\left\{ 
\begin{aligned}
&\Delta_{c_{y_i}} = 0, \quad if\,\, \sum_{j=1}^N \delta(y_j = y_i)=0,\\
&\Delta_{c_{y_i}} = \frac{\sum_{j=1}^N \delta(y_j = y_i)(c_{y_i} - F(x_i))}{\sum_{j=1}^N \delta(y_j = y_i)}, \quad otherwise.
\end{aligned}
\right. 
\end{equation}
where $\delta(\cdot) = 1$ if the condition inside $()$ holds true, and $\delta(\cdot) = 0$ otherwise. 

The classifier $C$ will discriminate the label of samples based on semantic features. It consists of several fully connected layers. Furthermore, cross entropy loss $L_{ce}$ is utilized to control the error of classifier $C$, which is defined as
\begin{equation}\label{ce}
     L_{ce} = - \frac{1}{N} \sum_{i=1}^N y_i \log(C(F(x_i)))
\end{equation}
where $C(F(x_i))$ is the prediction of $x_i$.

Further, auto-encoder \cite{chen2016variational,higgins2017beta,ng2011sparse} is used in order to retain the effective semantic information in $Z$. As shown in the right part of Fig \ref{2}, decoder $D$ is used to reconstruct $X$ from $Z$. It is made up of deconvolution, unpooling and fully connected layers. Among them, unpooling is the reverse of pooling and deconvolution is the reverse of convolution. Specifically, max unpooling keeps the maximum position information during max pooling, and then it restores the maximum values to the corresponding positions and set zeros to the rest positions as shown in Fig. \ref{3:a}. Analogously, average unpooling expands the feature map in the way of copying it as shown in Fig. \ref{3:b}. 

The deconvolution is also called transpose convolution to recover the shape of input from output, as shown in Fig. \ref{3:c}. See appendix A for the detailed convolution and deconvolution Operation, as well as toy examples. 

In addition, reconstruction loss is utilized to evaluate the difference between original signal data and reconstructed signal data. 
\begin{equation}\label{ae}
    L_{r} = \frac{1}{2N} \sum_{i=1}^N ||D(F(x_i)) - x_i||_2^2
\end{equation}
where $D(F(x_i))$ is the reconstruction of signal $x_i$. Intuitively, the more complete signal is reconstructed, the more valid information is carried within $Z$. Thus, the auto-encoder greatly helps the model to generate appropriate semantic features. 

As a result, the total loss function combines cross entropy loss, center loss and reconstruction loss as
\begin{equation}
    L_t = L_{ce} + \lambda_{ct}L_{ct} + \lambda_{r}L_{r}
\end{equation}
where the weights $\lambda_{ct}$ and $\lambda_{r}$ are used to balance the three loss functions. We have carefully designed the total loss function. The cross entropy loss is used to learn information from labels. And center loss minimizes the intra-class variations in the semantic space while keeping the inter-classes’ semantic features well separated, which also helps unknown classes to separate. Reconstruction loss makes model learn more information about signal data, because of well-reconstructed data. Ablation study in Section \ref{experiments} also validates the above points. The whole learning process with loss $L_t$ is summarized in Algorithm \ref{alg:training}, where $\theta_F$, $\theta_C$, $\theta_D$ denote the model parameters of the feature extractor $F$, the classifier $C$ and the decoder $D$, respectively.  

\begin{algorithm}[htb]
   \caption{Pseudocode for \deepmodel~Update}
   \label{alg:training}
\begin{algorithmic}
   \REQUIRE Labeled input and output set $\left\{(x_i, y_i)\right\}$ and hyperparameters $N, \eta, \alpha, \lambda_{ct}, \lambda_{r}$.
   \ENSURE Parameters $\theta_F, \theta_C, \theta_D$ and $\left\{c_j \right\}$.
   \STATE Initial parameters $\theta_F, \theta_C, \theta_D$.
   \STATE Initial parameter $\left\{c_j | j \in K \right\}$.
   \REPEAT
   \FOR{each batch with size $N$}
   \STATE Update $c_j$ for each $j$ : $c_j \leftarrow c_j - \alpha\Delta_{c_j}$
   \STATE Calculate $L_{ct}$ via Eq. \eqref{ct}. 
   \STATE Calculate $L_{ce}$ via Eq. \eqref{ce}.
   \STATE Calculate $L_{r}$ via Eq. \eqref{ae}.
   \STATE $L_t = L_{ce} + \lambda_{ct}L_{ct} + \lambda_{r}L_{r}$.
   \STATE Update $\theta_F$ : $\theta_F \leftarrow \theta_F - \eta\nabla _{\theta_F}L_t$.
   \STATE Update $\theta_C$ : $\theta_C \leftarrow \theta_C - \eta\nabla _{\theta_C}L_t$.
   \STATE Update $\theta_D$ : $\theta_D \leftarrow \theta_D - \eta\nabla _{\theta_D}L_t$.
   \ENDFOR
   \UNTIL{convergence}
\end{algorithmic}
\end{algorithm}

\subsection{Discriminator}
The discriminator $P$ is the tail but the core of the proposed framework. It discriminates among known and unknown classes based on the latent semantic space $Z$. For each known class $k$, the feature extractor $F$ extracts and computes the corresponding semantic center vector $S_k$ as:
\begin{equation}\label{si}
    S_k = \frac{\sum_{j=1}^{m} \delta(y_j = k)F(x_j)}{\sum_{j=1}^{m}\delta(y_j = k)}
\end{equation}
where $m$ is the number of all training samples. When a test signal $\mathcal{I}$ appears and $F(\mathcal{I})$ is obtained, the difference between the vector $F(\mathcal{I})$ and $S_k$ can be measured for each $k$. Specifically, the generalized distance between $F(\mathcal{I})$ and $S_k$ is used, which is defined as follows:
\begin{equation}\label{d}
    d(F(\mathcal{I}), S_k) = \sqrt{(F(\mathcal{I}) - S_k)^TA_k^{-1}(F(\mathcal{I}) - S_k)}
\end{equation}
where $A_k$ is the transformation matrix associated with class $k$ and $A_k^{-1}$ denotes the inverse of matrix $A_k$. When $A_k$ is the covariance matrix $\Sigma$ of semantic features of signals of class $k$, $d(\cdot, \cdot)$ is called Mahalanobis distance. When $A_k$ is the identity matrix\footnote{This is also the only possible choice in the case when the covariance matrix $\Sigma$ is not available, which happens for example when the signal set of some class is singleton. } $I$, $d(\cdot, \cdot)$ is reduced to Euclidean distance. $A_k$ also can be $\Lambda$ and $\sigma^2I$ where $\Lambda$ is a diagonal matrix formed by taking diagonal elements of $\Sigma$ and $\sigma^2\triangleq \frac{trace(\Sigma)}{t}$ with $t$ being the dimension of $S_k$. The corresponding distance based on $A_k=\Lambda$ and $A_k=\sigma^2I$ are called the second distance and third distance. Note that when the Mahalanobis distance, second distance and third distance are applied, the covariance matrix of each known class needs to be computed in advance.

With the above distance metric, we can establish our discriminant model which is divided into two steps. Firstly, distinguish between known and unknown classes. Secondly, discriminate which known classes or unknown classes the test signal belongs to. \emph{The first step} is done by comparing the threshold $\Theta_1$ with the minimal distance $d_1$ given by
\begin{equation}\label{d1}
    d_1 = \min_{S_k \in S} d(F(\mathcal{I}), S_k)
\end{equation}
where $S$ is the set of known semantic center vectors. Let us denote by $y_{\mathcal{I}}$  the prediction of $\mathcal{I}$. If $ d_1< \Theta_1$, $y_{\mathcal{I}} \in K$, otherwise $y_{\mathcal{I}} \in U$. Owing to utilizing the center loss in training, the semantic features of signals of class $k$ are assumed to obey multivariate Gaussian distribution.
Inspired by the three-sigma rule \cite{pukelsheim1994three}, we set $\Theta_1$ as follows
\begin{equation} \label{3sigma}
    \Theta_1 = \lambda_1 \times 3\sqrt{t}
\end{equation} 
where $\lambda_1$ is a control parameter referred to as the \textit{discrimination coefficient}. 

Two remarks are made as follows to explain the Gaussian distribution assumption and the choice of $\Theta_1$, respectively.
\begin{remark}
\emph{
In our loss function, we have the center loss component which aims to minimize \eqref{ct} with respect to the semantic layer. It is not difficult to show that
\begin{align}
\begin{split}
&\arg \min_{\theta_F} L_{ct} = \arg \max_{\theta_F} -L_{ct}\\
&=\arg \max_{\theta_F} -\frac{1}{2N}\sum_{i=1}^N ||F(x_i) - c_{y_i}||^2_2\\
&=\arg \max_{\theta_F} -\frac{1}{2N}\sum_{i=1}^N (F(x_i) - c_{y_i})^T(F(x_i) - c_{y_i})
\end{split}
\end{align}  
Because of the monotonicity of exponential function, we have
\begin{align}
\begin{split}
&\arg \max_{\theta_F} -\frac{1}{2N}\sum_{i=1}^N (F(x_i) - c_{y_i})^T(F(x_i) - c_{y_i})\\
&=\arg \max_{\theta_F} e^\frac{1}{N} \prod_{i=1}^N e^{-\frac{(F(x_i) - c_{y_i})^T(F(x_i) - c_{y_i})}{2}}\\
&=\arg \max_{\theta_F} e^\frac{1}{N} \prod_{i=1}^N e^{-\frac{(F(x_i) - c_{y_i})^TI^{-1}(F(x_i) - c_{y_i})}{2}}\\
&=\arg \max_{\theta_F} e^\frac{1}{N} (2\pi)^{\frac{t}{2}}|I|^{\frac{1}{2}}\\
&\qquad \qquad \prod_{i=1}^N \frac{1}{(2\pi)^{\frac{t}{2}}}\frac{1}{|I|^{\frac{1}{2}}} e^{-\frac{(F(x_i) - c_{y_i})^TI^{-1}(F(x_i) - c_{y_i})}{2}}
\end{split}
\end{align} 
where $t$ denotes the dimension of Gaussian distribution and $I$ denotes the identity matrix. Let $\beta \triangleq e^\frac{1}{N} (2\pi)^{\frac{t}{2}}|I|^{\frac{1}{2}}$ and $P(F(x_i)|y_i) = \frac{1}{(2\pi)^{\frac{t}{2}}}\frac{1}{|I|^{\frac{1}{2}}}e^{-\frac{(F(x_i) - c_{y_i})^TI^{-1}(F(x_i) - c_{y_i})}{2}}$, the above equation can be equivalently written as
\begin{equation}
   \arg \max_{\theta_F} \beta \prod_{i=1}^N P(F(x_i)|y_i)
\end{equation}
where $P(F(x_i)|y_i) = \mathcal N (c_{y_i}, I)$. This indicates that very likely the output of the semantic layer follows the Gaussian distribution.\footnote{Note that, however, due to the existence of the other two component loss functions, we propose using a general covariance matrix to describe the output of the semantic layer, as shown in \eqref{d}. }
}
\end{remark}

\begin{remark}
\emph{
The choice of $\Theta_1$ in \eqref{3sigma} is made due to the following two considerations. First, the well-known three-sigma rule of thumb is often used for identification of outliers \cite{bajorski2011statistics}. It is shown in \cite{bajorski2011statistics} that this rule should be properly generalized due to the impact of the dimension in the mult-dimensional case. We here present a natural generalization to the t-dimensional case by simply averaging the Mahalanobis distance over $\sqrt{t}$, so as to remove the impact of the dimension on the choice of $\Theta_1$.  The above explains why we have the term $3\sqrt{t}$ in \eqref{3sigma}. Second, a control parameter $\lambda_1$  is incorporated to make the choice of $\Theta_1$ more sophisticated so that it can work well for  complex recognition tasks. Our numerical experiments later validate the effectiveness of the choice of $\Theta_1$.   
}
\end{remark}

\emph{The second step} is more complicated. If $\mathcal{I}$ belongs to the known classes, its label $y_{\mathcal{I}}$ can be easily obtained via
\begin{equation} \label{s}
    y_{\mathcal{I}} = \arg \min_{k} d(F(\mathcal{I}), S_k).
\end{equation}
Obviously the main difficulty lies in dealing with the case when  $\mathcal{I}$ is classified as unknown in the first step. To illustrate, let us denote by $R$ the recorded unknown classes and define $S_R$ to be the set of the semantic center vectors of $R$. In this difficult case with $R \subseteq \varnothing$, a new signal label $R_1$ is added to $R$ and $F(\mathcal{I})$ is set to be the semantic center vector $S_{R_1}$. The unknown signal $\mathcal{I}$ is saved in set $G_{R_1}$ and let $y_{\mathcal{I}} = R_1$. While in the difficult case with $R \not\subseteq \varnothing$, the threshold $\Theta_2$ is compared to the minimal distance $d_2$ which is defined by
\begin{equation}\label{d2}
    d_2 = \min_{S_{R_u} \in S_R} d(F(\mathcal{I}), {R_u}) 
\end{equation}

\begin{table*}[htbp]
\caption{Standard metadata of dataset 2016.10A. For a larger version, 2016.10B, the class "AM-SSB" is removed, while the number of samples for each class is sixfold (120000). For a smaller one, 2016.04C, all 11 classes is included, but the number of samples for each class is disparate (range from 4120 to 24940).}
\centerline{\begin{tabular}{|c|c|c|c|c|} 
 \hline
 total samples & \# of samples each class & \# of samples each SNR & feature dimension & classes (modulations) \\ [0.5ex]
 \hline
 220000 & 20000 & 1000 & $2\times128$ & 11 \\ 
 \hline
\multicolumn{5}{|c|}{modulation types}\\
\hline
\multicolumn{5}{|c|}{8PSK, AM-DSB, AM-SSB, BPSK, CPFSK, GFSK, PAM4, QAM16, QAM64, QPSK, WBFM}\\
\hline
\multicolumn{2}{|c}{\# of SNR values}&
\multicolumn{3}{|c|}{SNR values}\\
\hline
\multicolumn{2}{|c}{20}&
\multicolumn{3}{|c|}{-20,-18,-16,-14,-12,-10,-8,-6,-4,-2,0,2,4,6,8,10,12,14,16,18}\\
\hline
\end{tabular}}
\label{table:dataset}
\end{table*}

Intuitively, a good choice of $\Theta_2$ may be made based on the distance between $F(x)$ and $S_k$'s. $d_1$ is the minimum distance which is firstly used in our test of choice of $\Theta_2$. Actually, we test a set of choices of $\Theta_2$ and numerically find that unknown classes can be often correctly identified when  $\Theta_2$ is set between $d_1$ and $d_{med}$, where $d_{med}$ is the median distance between $F(x)$ and each $S_k$. Therefore, the threshold $\Theta_2$ is finally set as 
\begin{equation}\label{theta2}
    \Theta_2 =  \frac{d_1 + \lambda_2 \times d_{med}}{1 + \lambda_2}
\end{equation}
where $\lambda_2$ is used to balance the two distances $d_1$ and $d_{med}$.

To proceed, let $n_R$ denote the number of recorded signal labels in $R$. Then, if $d_2>\Theta_2$, a new signal label $R_{n_R+1}$ is added to $R$ and set $y_{\mathcal{I}} = n_R+1$. Note that we don't impose any prior restrictions on the value of $n_R$ (the size of set $R$), i.e., our model can never know the number of the unknown classes pending to be discriminated. 
Then if $d_2\leq\Theta_2$, we set 
\begin{equation} \label{rs}
    y_{\mathcal{I}} = \arg \min_{R_u} d(F(\mathcal{I}), S_{R_u}).
\end{equation}
and save the signal $\mathcal{I}$ in $G_{y_{\mathcal{I}}}$. Accordingly, $S_{y_{\mathcal{I}}}$ is updated via
\begin{equation}\label{upd}
    S_{y_{\mathcal{I}}} = \frac{\sum_{k \in G_{y_{\mathcal{I}}}} F(k)}{\#(G_{y_{\mathcal{I}}})}
\end{equation}
where $\#(G_{y_{\mathcal{I}}})$ denotes the number of signals in set $G_{y_{\mathcal{I}}}$. As a result, with the increase of the number of predictions for unknown signals, the model will gradually improve itself by way of refining $S_{R_u}$'s. 

\begin{algorithm}[h]
  \caption{Pseudocode for Discriminator $P$}
  \label{alg:2}
\begin{algorithmic}
  \REQUIRE Test input $\left\{(\mathcal{I})\right\}$, transformation matrices $\left\{A_k, A_{R_u}\right\}$, sets $S, R, S_R, D$ and hyperparametes $\Theta_1$, $\Theta_2$.
  \ENSURE $y_{\mathcal{I}}$.
  \STATE Calculate $F(\mathcal{I})$.
  \STATE Calculate $d_1$ via Eq. \eqref{d1}.
  \STATE Calculate $d_2$ via Eq. \eqref{d2}.
  \IF{$d_1 < \Theta_1 $} 
  \STATE Calculate $y_{\mathcal{I}}$ via Eq. \eqref{s}.
  \ELSIF{$d_1 \geq \Theta_1$ and $R \subseteq \varnothing$}
  \STATE Add $R_1$ to $R$.
  \STATE $y_{\mathcal{I}} = R_1$ .
  \ELSIF{$d_1 \geq \Theta_1$, $R \not\subseteq \varnothing$ and $d_2 > \Theta_2$}
  \STATE Add $R_{n_R+1}$ to $R$.
  \STATE $y_{\mathcal{I}} = R_{n_R+1}$.
  \ELSE
  \STATE Calculate $y_{\mathcal{I}}$ via Eq. \eqref{rs}
  \ENDIF 
  \STATE Save $\mathcal{I}$ in $G_{y_{\mathcal{I}}}$.
  \STATE update $S_{y_{\mathcal{I}}}$ via Eq. \eqref{upd}.
\end{algorithmic}
\end{algorithm}

\begin{figure*}[h!]
\centerline{\includegraphics[width=19cm]{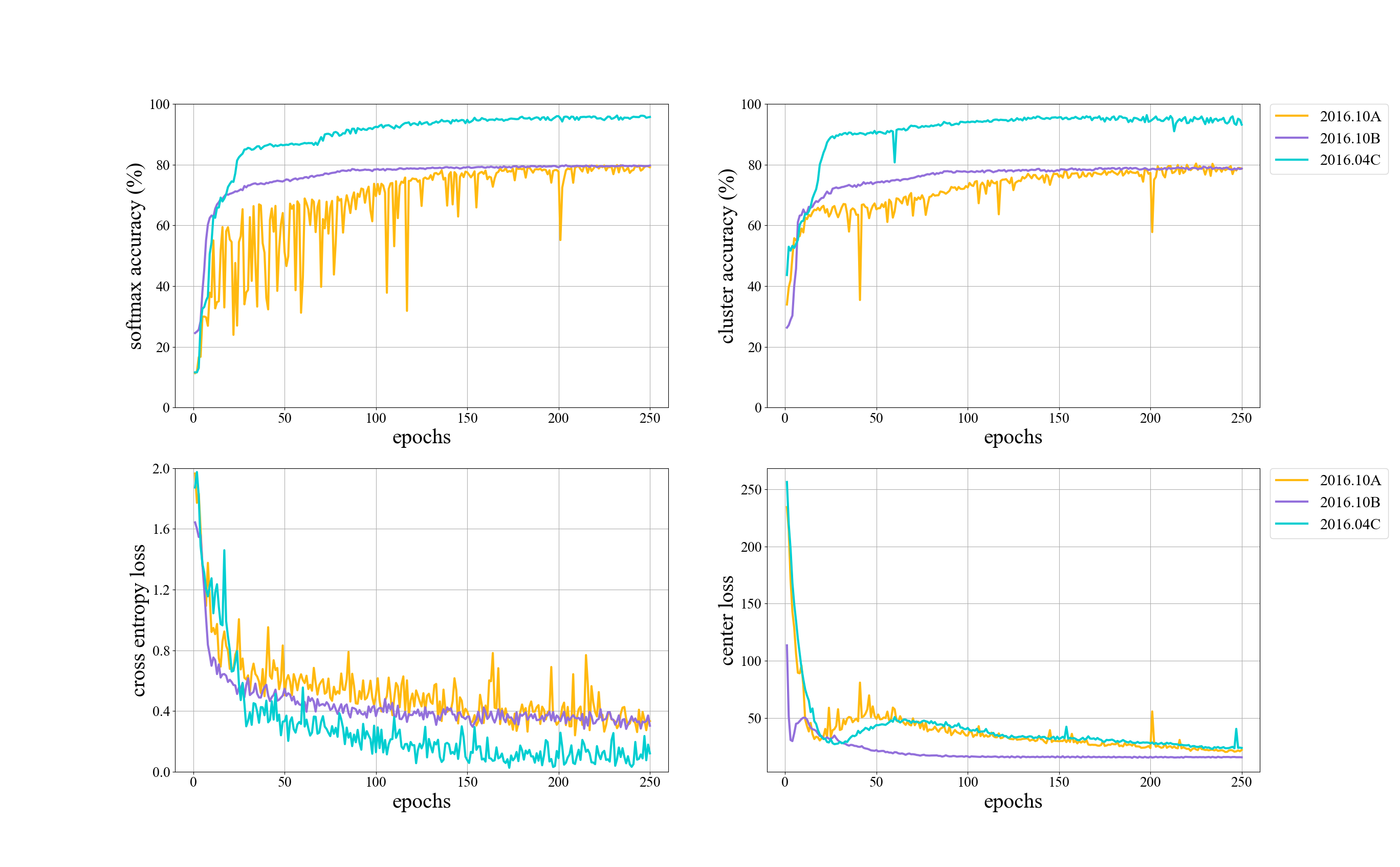}}
\caption{In-training statistics on three datasets. The accuracy is based on the known test set.}
\centering
\label{figure:intraining}
\end{figure*}

To summarize, we present the whole procedure of the discriminator in Algorithm \ref{alg:2}. We emphasize that our SR2CNN is different from the common open-set recognition methods. Assuming that there are $n$ known classes and an uncertain number of unknown classes, the traditional open-set recognition method will only distinguish the test samples into $n+1$ classes, while SR2CNN will distinguish the test samples into $n+n_R$ classes via Algorithm \ref{alg:2}, where $n_R$ is the number of unknown classes recognized by the discriminator. Specifically, for the case when a test sample belongs to an unknown class, we determine whether it belongs to an existing unknown class or a new unknown class by comparing $d_2$ with threshold $\Theta_2$. Hence, the notable advantage of SR2CNN over the common open-set recognition method lies in that SR2CNN can roughly distinguish how many unknown classes there are in the test set, not just label the test sample as unknown.

\section{Experiments and Results}
\label{experiments}
\begin{table*}[h!]
{
\caption{Contrast between supervised learning and our \experiment~learning scenario on three datasets. Dash lines in the \experiment~column specify the boundary between known and unknown classes. \textbf{Bold}: accuracy for a certain unknown class. \textit{Italic}: accuracy computed only to help draw a transverse comparison. Items split by slash "/" like "75.9\%/8.4\%" denote the accuracy respectively for two isotopic classes. “-” denotes no corresponding result for such case.}
\label{table:2}
}
\centerline{\begin{tabular}{|c|c||c|c||c|c||c|c|} 
 \hline
 \multicolumn{2}{|c||}{\multirow{2}*{\diagbox{indicator}{scenario}}} & \multicolumn{2}{c||}{2016.10A} & \multicolumn{2}{c||}{2016.10B} &
 \multicolumn{2}{c|}{2016.04C}\\
 \cline{3-8}
\multicolumn{2}{|c||}{} & supervised & \experiment & supervised & \experiment & supervised & \experiment\\
\hline\hline
\multirow{12}{4em}{accuracy} & 8PSK (1) & 85.0\% & 85.5\% & 95.5\% & 86.7\% & 74.9\% & 69.3\%\\
\cline{2-2}
 ~ & AM-DSB (2) & 100.0\% & 73.5\% & 100.0\% & 41.3\% & 100.0\% & 91.1\%\\
\cline{2-2}
 ~ & BPSK (4) & 99.0\% & 95.0\% & 99.8\% & 96.5\% & 99.8\% & 97.6\%\\
\cline{2-2}
 ~ & PAM4 (7) & 98.5\% & 94.5\% & 97.6\% & 93.4\% & 99.6\% & 96.8\%\\
\cline{2-2}
 ~ & QAM16 (8) & 41.6\% & 49.3\% & 56.8\% & 40.0\% & 97.6\% & 98.4\%\\
\cline{2-2}
 ~ & QAM64 (9) & 60.6\% & 44.0\% & 47.5\% & 49.6\% & 94.0\% & 97.6\%\\
\cline{2-2}
 ~ & QPSK (10) & 95.0\% & 90.5\% & 98.9\% & 90.6\% & 86.8\% & 81.5\%\\
\cline{2-2}
 ~ & WBFM (11) & 38.2\% & 32.0\% & 39.6\% & 50.4\% & 88.8\% & 86.9\%\\
\cline{2-2}\cdashline{6-6}
~ & \textbf{CPFSK (5)} & 100.0\% & 99.0\% & 100.0\% & \textbf{75.9\%/8.4\%} & 100.0\% & 96.2\%\\
\cline{2-2}\cdashline{4-4}\cdashline{8-8}
 ~ & \textbf{GFSK (6)} & 100.0\% & \textbf{99.0\%} & 100.0\% & \textbf{95.6\%/2.3\%} & 100.0\% & \textbf{82.0\%}\\
\cline{2-2}
  ~ & \textbf{AM-SSB (3)} & 100.0\% & \textbf{100.0\%} & - & - & 100.0\% & \textbf{100.0\%}\\
 \hline\hline
 \multicolumn{2}{|c||}{total accuracy} & 83.5\% & \textit{78.4\%} & 83.6\% & \textit{72.0\%} & 94.7\% & \textit{91.5\%}\\
 \hline
 \multicolumn{2}{|c||}{average known accuracy} & \textit{79.8\%} & 73.7\% & \textit{79.5\%} & 68.5\% & \textit{93.5\%} & 91.6\%\\
 \hline
 \multicolumn{2}{|c||}{true known rate} & - & 95.9\% & - & 86.9\% & - & 97.0\%\\
 \hline
 \multicolumn{2}{|c||}{true unknown rate} & - & 99.5\% & - & 91.1\% & - & 90.0\%\\
 \hline
\end{tabular}}
\end{table*}

\begin{figure*}[htp]
\centerline{\includegraphics[width=19cm]{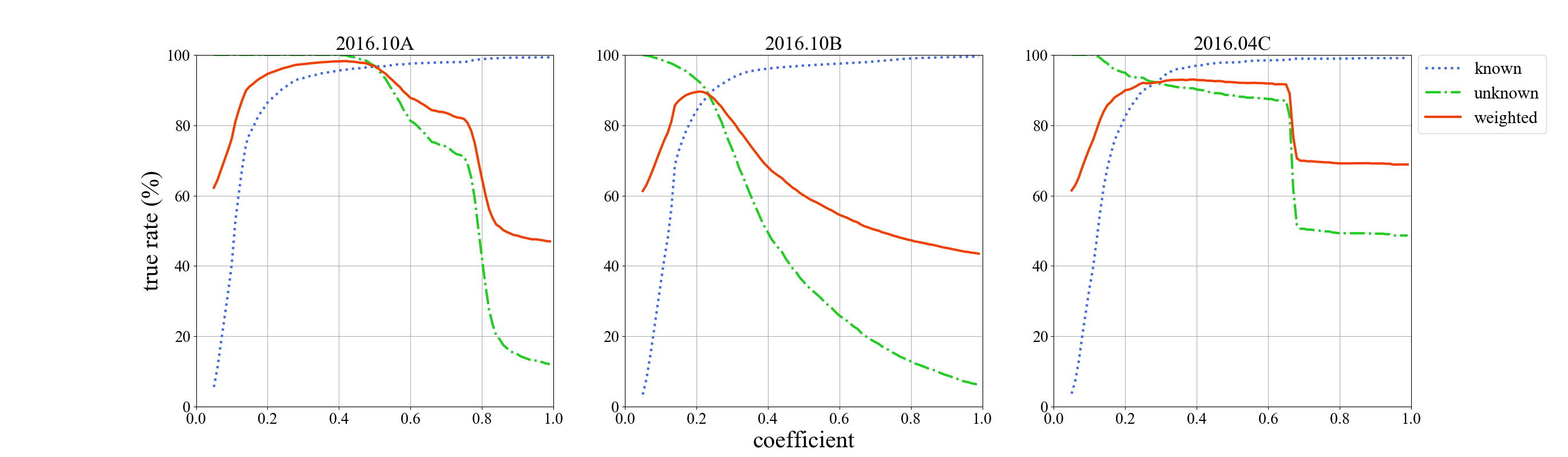}}
\caption{Correlation between true known/unknown accuracy and discrimination coefficient ($\lambda_1$) on three datasets.}
\centering
\label{figure:2}
\end{figure*}

In this section, we demonstrate the effectiveness of the proposed \deepmodel~approach by conducting extensive experiments with the dataset 2016.10A, as well as its two counterparts, 2016.10B and 2016.04C \cite{o2016convolutional}. The data description is presented in Table \ref{table:dataset}. All $11$ types of modulations are numbered with class labels from left to right.

\textbf{Sieve samples.} Samples with SNR less than 16 are firstly filtered out, only leaving a purer and higher-quality portion (one-tenth of origin) to serve as the overall datasets in our experiments.

\textbf{Choose unknown classes.} Empirically, a class whose features are hard to learn is an arduous challenge for a standard supervised learning model, let alone when it plays an unknown role in our \experiment~scenario (especially when no prior knowledge about the number of the unknown classes is given, as we mentioned in the Subsection 4.2). Hence, necessarily, a completely supervised learning stage is carried out beforehand, to help us nominate suitable unknown classes. If the prediction accuracy of the full supervision method is rather low for certain class, it is reasonable to exclude this class in ZSL, because ZSL will definitely not yield a good performance for this class. In our experiments, unknown classes are randomly selected from a set of classes for which the accuracy of full supervision is higher than 50\%. As shown in Table \ref{table:2}, the ultimate candidates fall on AM-SSB(3) and GFSK(6) for 2016.10A and 2016.04C, while CPFSK(5) and GFSK(6) for 2016.10B.

\textbf{Split training, validation and test data.} 70\% of the samples from the known classes make up the overall training set while 15\% makes up the known validation set and the rest 15\% makes up the known test set. For the unknown classes, there's only a test set needed, which consists of 15\% of the unknown samples.

Due to the three preprocessing steps, we get a small copy of, e.g., dataset 2016.10A, which contains a training set of $12600$ samples, a known validation set of $2700$ samples, a known test set of $2700$ samples and an unknown test set of $600$ samples.

All of the networks in \deepmodel~are computed on a single GTX Titan X graphic processor and implemented in Python, and trained using the Adam optimizer with learning rate $\eta= 0.001$ and batch size $N=256$. Generally, we allow our model to learn and update itself maximally for 250 epochs. In addition, the grid search is applied to the validation set to determine the hyperparameters.

\subsection{In-training Views}
Basically, the average softmax accuracy of the known test set will converge roughly to  $80\%$ on both 2016.10A and 2016.10B, while to $94\%$ on 2016.04C, as indicated in Fig. \ref{figure:intraining}. Note that there's almost no perceptible loss on the accuracy when using the clustering approach (i.e., the distance measure-based classification method described in Section IV) to predict instead of softmax, meaning that the semantic feature space established by our \deepmodel~functions very well. For ease of exposition, we will refer to the \textit{known cluster accuracy} as \textit{upbound (UB)}. 

During the training course, the cross entropy loss undergoes sharp and violent oscillations. This phenomenon makes sense, since the extra center loss and reconstruction loss will intermittently shift the learning focus of the \deepmodel. 

\begin{table*}[htb]
{
\caption{Ablation study about the discrimination task via $P$ on 2016.10A in test. \textbf{Bold}: performance of the original \deepmodel~model. F1 score denotes $2 \times accuracy \times precision / (accuracy + precision)$}
\label{table:6}
}
\centerline{\begin{tabular}{|c|c|c|c|c|c|c|} 
 \hline
\multicolumn{2}{|c|}{\diagbox{indicator}{modification}} & \deepmodel & without Cross Entropy Loss & without Center Loss & without Reconstruction Loss & L1 Loss\\
\hline
\multirow{2}*{accuracy} & AM-SSB(3) & \textbf{100.0\%} & 100.0\% & 99.5\% & 100.0\% & 100.0\%\\
\cline{2-2}
~ & GFSK(6) & \textbf{99.5\%} & 98.5\% & 61.0\% & 94.8\% & 95.8\% \\
\hline\hline
\multicolumn{2}{|c|}{average known accuracy} & \textbf{73.7\%} & 72.1\% & 69.0\% & 72.3\% & 70.4\% \\
\hline
\multirow{2}*{precision} & known & \textbf{76.8\%} & 75.3\% & 79.1\% & 74.5\% & 82.8\% \\
\cline{2-7}
~ & unknown & \textbf{96.1\%} & 95.2\% & 82.4\% & 94.5\% & 86.1\% \\
\hline
\multirow{2}*{F1 score} & known & \textbf{75.3\%} & 73.6\% & 73.7\% & 73.3\% & 76.1\% \\
\cline{2-7}
~ & unknown & \textbf{98.0\%} & 97.2\% & 81.3\% & 95.9\% & 91.6\% \\
\hline
\end{tabular}}
\end{table*}

\subsection{Critical Results}
The most critical results are presented in Table \ref{table:2}.
To better illustrate it, we will firstly make a few definitions in analogy to the binary classification problem. By superseding the binary condition \textit{positive} and \textit{negative} with \textit{known} and \textit{unknown} respectively, we can similarly elicit \textit{true known (TK)}, \textit{true unknown (TU)}, \textit{false known (FK)} and \textit{false unknown (FU)}. Subsequently, we get two important indicators as follows:
\[true~known~rate~(TKR)~=\frac{TK}{K}=\frac{TK}{TK+FU}\]
\[true~unknown~rate~(TUR)~=\frac{TU}{U}=\frac{TU}{TU+FK}\]
Furthermore, we define precision likewise as follows:
\[known~precision~(KP)=\frac{S_{correct}}{TK+FK}\]
\[unknown~precision~(UP)=\frac{U_{dominantly\_correct}}{TU+FU}\]
where $S_{correct}$ denotes the total number of known samples that are classified to their exact known classes correctly. $U_{dominantly\_correct}$ denotes the total number of unknown samples that are classified to their exact newly-identified unknown classes correctly. For evaluation, the real label of a certain newly-recorded unknown class is determined as the label of the most signal samples in that class. Note that sometimes unexpectedly, our \deepmodel~may classify a small portion of signals into different unknown classes but their real labels are actually identical and correspond to one certain unknown class (we name these unknown classes as isotopic classes). In this rare case, we only count the identified unknown class with the highest accuracy in calculating $U_{dominantly\_correct}$.

For \experiment, we test our \deepmodel~with several different combinations of aforementioned parameters $\lambda_1$ and $\lambda_2$, hopefully to snatch a certain satisfying result out of multiple trials. Fixing $\lambda_2$ to 1 simply leads to fair performance, though still, we adjust $\lambda_1$ in a range between 0.05 and 1.0. Here, the pre-defined indicators above play an indispensable part to help us sift the results. Generally, a well-chosen result is supposed to meet the following requirements: \textbf{1.} the \textbf{weighted true rate (WTR)}: ~\textit{$0.4\times$TKR+$0.6\times$TUR} ~is as great as possible; \textbf{2.} \textit{KP$>0.95\times$UB}, where \textit{UB} is the \textit{upbound} defined as the \textit{known cluster accuracy}; \textbf{3.} \textit{$\#^j_{isotopic}<=$2} for all possible $j$, where $\#^j_{isotopic}$ denotes the number of isotopic classes corresponding to a certain unknown class $j$.\par

\begin{figure}[h]
\includegraphics[width=9cm]{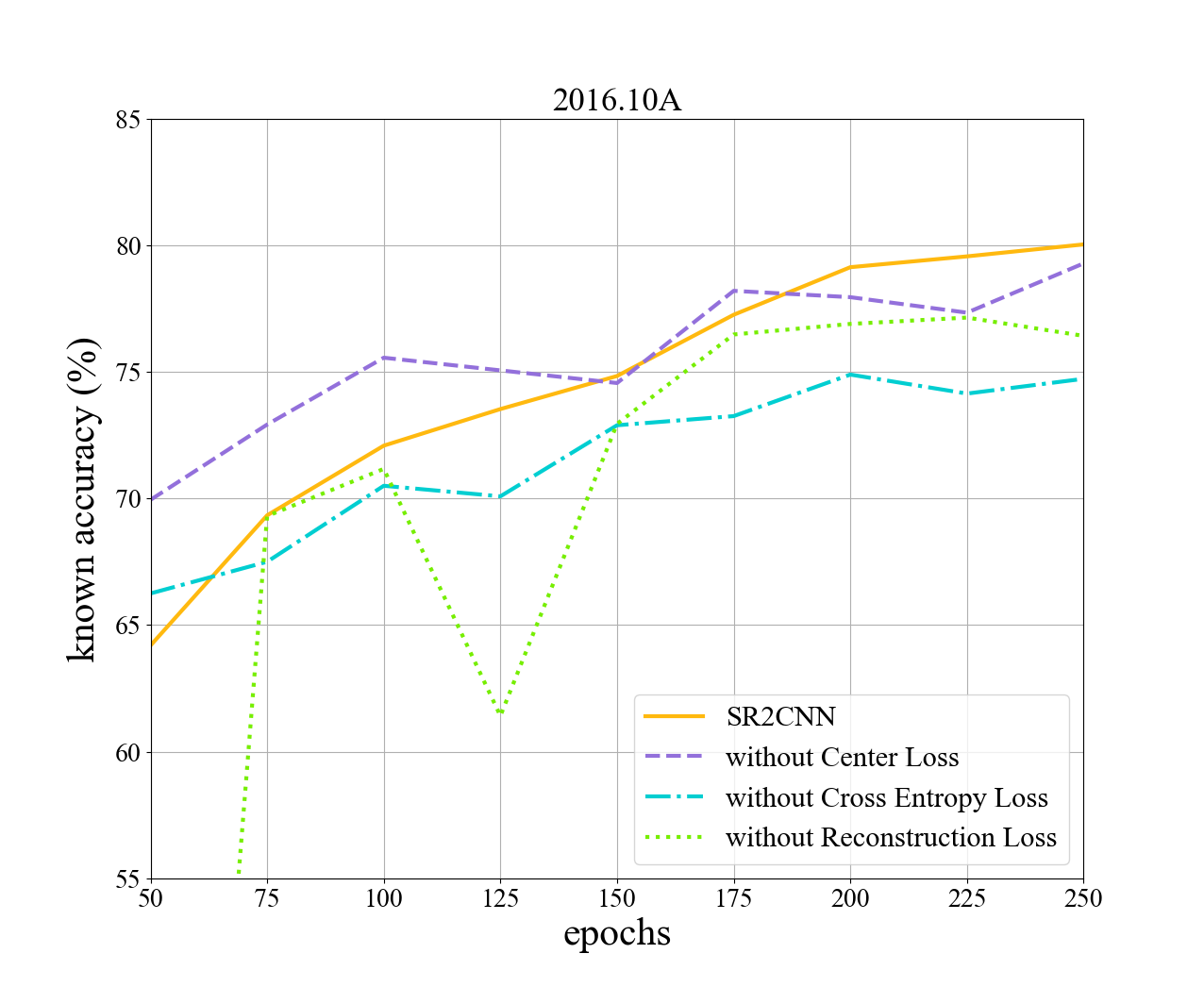}
\caption{Ablation study about the known accuracy on 2016.10A in training.}
\centering
\label{figure:4}
\end{figure}

\begin{figure}[h]
\includegraphics[width=9cm]{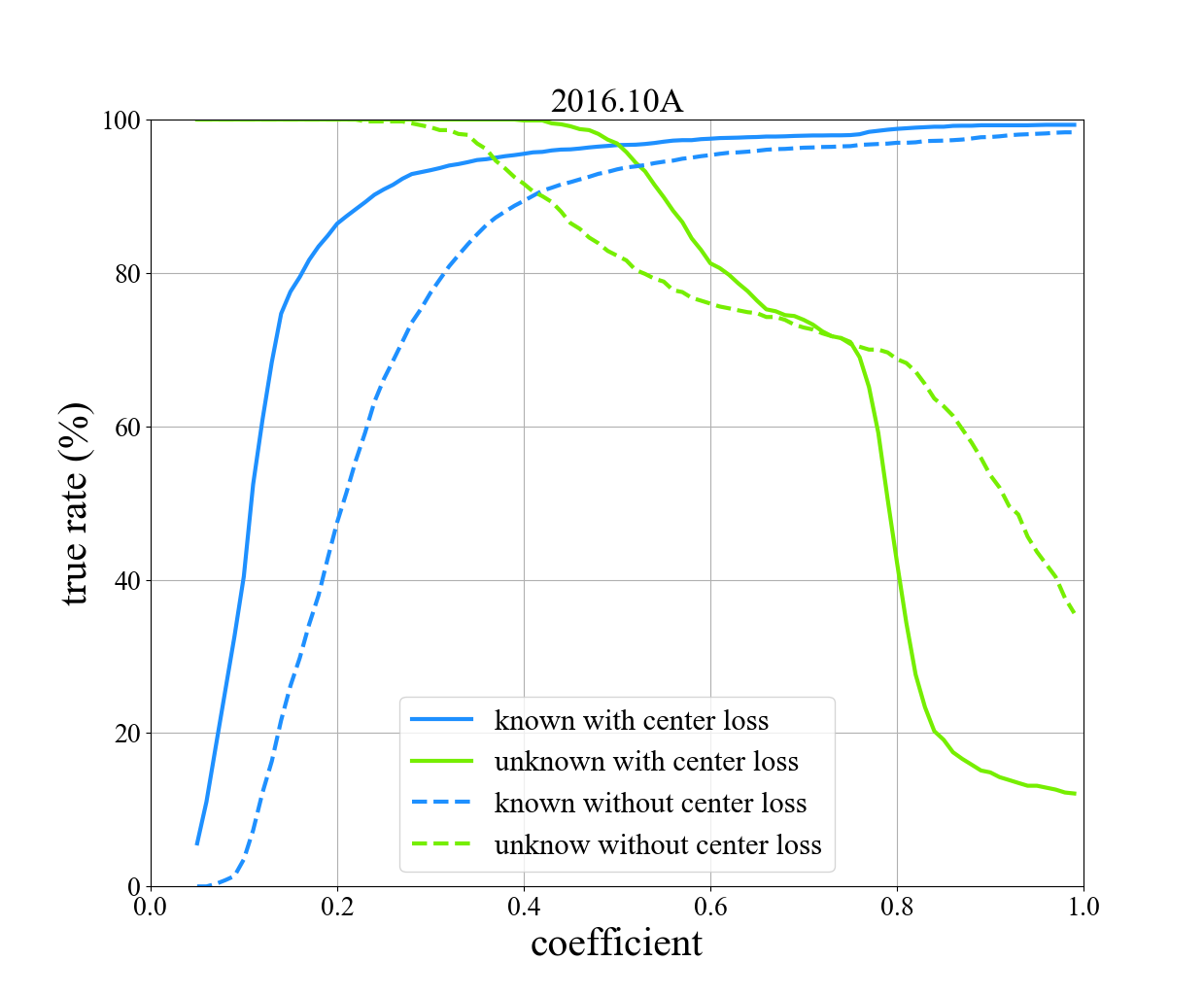}
\caption{Effect of center loss. The presence of center loss is distinguished by line shape(solid or dash). Interviewees(known or unknown accuracy) are distinguished by line color(blue or green).}
\centering
\label{figure:3}
\end{figure}

\begin{table*}[htb]
{
\caption{Performance among different set of chosen unknown classes on 2016.10A. \textbf{Bold}: recall rate. Item split by slash "/" like "87.8\%/9.0\%" and "-" basically are of the same meanings with Table \ref{table:2}.}
\label{table:3}
}
\centerline{\begin{tabular}{|c|c|c|c|c|c|} 
 \hline
 \multicolumn{2}{|c|}{\multirow{2}*{\diagbox{indicator}{training config}}} & \multicolumn{4}{c|}{unknown classes}\\
\cline{3-6}
\multicolumn{2}{|c|}{} & AM-SSB and GFSK & CPFSK and GFSK & AM-SSB and CPFSK & AM-SSB, CPFSK and GFSK \\
\hline
\multirow{3}{4em}{accuracy} & AM-SSB(3) & 100.0\% & - & 100.0\% & 100.0\%\\
\cline{2-2}
~ & CPFSK(5) & - & 71.0\% & 87.8\%/9.0\% & 65.5\%\\
\cline{2-2}
~ & GFSK(6) & 99.5\% & 100.0\% & - & 90.5\%\\
\hline\hline
\multicolumn{2}{|c|}{average known accuracy} & 73.7\% & 68.3\% & 75.6\% & 69.6\%\\
\hline
\multicolumn{2}{|c|}{true known rate} & 95.9\% & 89.6\% & 96.2\% & 90.9\%\\
\hline
\multicolumn{2}{|c|}{true unknown rate} & 99.8\% & 85.5\% & 98.4\% & 85.4\%\\
\hline\hline
\multirow{2}*{precision} & known & \textbf{76.8\%} & \textbf{73.6\%} & \textbf{78.3\%} & \textbf{74.0\%}\\
\cline{2-6}
~ & unknown & \textbf{96.1\%} & \textbf{89.2\%} & \textbf{91.9\%} & \textbf{90.4\%}\\
\hline
\end{tabular}}
\end{table*}

In order to better make a transverse comparision, we compute two extra indicators, average total accuracy in \experiment~scenario and also average known accuracy in completely supervised learning, shown as italics in Table \ref{table:2}. On the whole, the results are promising and excellent. However, we have to admit that \experiment~learning somewhat incurs a little bit performance loss as compared with the fully supervised model. Looking vertically, among all modulations, the performance loss especially occurs in the class AM-DSB. While looking horizontally among all datasets, the performance loss especially occurs in dataset 2016.10B. After all, when losing sight of the two unknown classes, \deepmodel~can only acquire a segment of the intact knowledge that shall be totally learned in a supervised case. It is this imperfection that presumably leads to an apparent variation on each class's accuracy when compared with supervised learning. Among these classes, the poorest victim is always AM-DSB, with considerable portion of its samples rejected as unknown ones. Besides, the features, especially those of the unknown classes, among these three datasets are not exactly in the same difficulty levels of learning. Some unknown features may even be similar to those known ones, which can consequently cause confusions in the discrimination tasks. It is no doubt that these uncertainties and differences in the feature domain matter a lot. Take 2016.10B, compared with its two counterparts, it emanates the greatest loss (more than 10\%) on average accuracy (both total and known), and also a pair of inferior true rates. Moreover, it is indeed the single case, where both two unknown classes are separately identified into two isotopic classes.

It is obvious that average accuracy strongly depends on the weighted true rate (WTR). Since the clearer for the discrimination between known and unknown, the more accurate for the further classification and identification. Therefore, to better study this discrimination ability, we depict Fig. \ref{figure:2} to elucidate its variation trends regarding discrimination coefficient ($\lambda_1$). At the same time, we introduce a new concept \textit{discrimination interval} as an interval where the weighted true rate is always greater than 80\%. The width of the above interval is used to help quantify this discrimination ability.  

\begin{table*}[htb]
{
\caption{Comparison between our \deepmodel~model and several traditional open-set model and outlier detectors on 2016.10A. \textbf{Bold}: performance of the dominant \deepmodel~model. \textit{Italic}: performance of these traditional methods when true known rates reach the highest. Vertical bar "$|$" is used to split the standard results and the italic ones.}
\label{table:4}
}
\scriptsize{
\centerline{\begin{tabular}{|c|c|c|c|c|c|c|c|} 
 \hline
 \diagbox{indicator}{detector} & \deepmodel & IsolationForest \cite{liu2008isolation} & EllipticEnvelope \cite{rousseeuw1999fast} & OneClassSVM \cite{chen2001one} & LocalOutlierFactor \cite{breunig2000lof} & OpenMax \cite{bendale2016towards} & MDL4OW \cite{liu2020few}\\
\hline
AM-SSB(3) & \textbf{100.0\%} & 72.3\% $|$ \textit{00.0\%} & 100.0\% $|$ \textit{100.0\%} & 96.3\% $|$ \textit{26.0\%} & 100.0\% & 100.0\% & 99.3\% \\
\cline{1-1}
GFSK(6) & \textbf{99.5\%} & 01.3\% $|$ \textit{00.0\%} & 90.0\% $|$ \textit{00.0\%} & 00.0\% $|$ \textit{00.0\%} & 00.0\% & 00.0\% & 26.5\% \\
\hline\hline
 true known rate & \textbf{95.9\%} & 81.3\% $|$ \textit{99.9\%} & 46.1\% $|$ \textit{97.6\%} & 85.5\% $|$ \textit{92.0\%} & 96.7\% & 98.1\% & 79.4\% \\
\hline
 true unknown rate & \textbf{99.8\%} & 36.8\% $|$ \textit{00.0\%} & 95.0\% $|$ \textit{50.0\%} & 48.1\% $|$ \textit{13.0\%} & 50.0\% & 50.0\% & 62.9\% \\
\hline
\end{tabular}}}
\end{table*}


\begin{table*}[h]
{
\caption{Contrast between supervised learning and our \experiment~learning scenario on dataset SIGNAL-202002. Dash lines in the \experiment~column specify the boundary between known and unknown classes. \textbf{Bold}: accuracy for a certain unknown class. \textit{Italic}: accuracy computed only to help draw a transverse comparision. "-" basically is of the same meanings with Table \ref{table:2}.}
\label{table:5}
}
\centerline{\begin{tabular}{|c|c||c|c|} 
 \hline
 \multicolumn{2}{|c||}{\multirow{2}*{\diagbox{indicator}{scenario}}} & \multicolumn{2}{c|}{SIGNAL-202002}\\
 \cline{3-4}
\multicolumn{2}{|c||}{} & supervised learning & zero-shot learning\\
\hline\hline
\multirow{11}{4em}{accuracy} & BPSK (1) & 84.3\% & 70.8\% \\
\cline{2-2}
 ~ & QPSK (2) & 86.5\% & 67.8\% \\
\cline{2-2}
 ~ & 8PSK (3) & 67.8\% & 70.3\% \\
\cline{2-2}
 ~ & 16QAM (4) & 99.5\% & 96.8\% \\
\cline{2-2}
 ~ & 64QAM (5) & 95.5\% & 84.8\% \\
\cline{2-2}
 ~ & PAM4 (6) & 97.0\% & 89.0\% \\
\cline{2-2}
 ~ & GFSK (7) & 56.3\% & 38.3\% \\
\cline{2-2}
 ~ & AM-DSB (10) & 63.8\% & 67.3\% \\
\cline{2-2}
 ~ & AM-SSB (11) & 44.3\% & 62.0\% \\
\cline{2-2}\cdashline{4-4}
 ~ & \textbf{CPFSK (8)} & 100.0\% & \textbf{81.0\%} \\
\cline{2-2}
 ~ & \textbf{B-FM (9)} & 93.5\% & \textbf{74.5\%} \\
 \hline\hline
 \multicolumn{2}{|c||}{average total accuracy} & 80.8\% & \textit{73.0\%} \\
 \hline
 \multicolumn{2}{|c||}{average known accuracy} & \textit{77.3\%} & 71.9\% \\
 \hline
 \multicolumn{2}{|c||}{true known rate} & - & 82.3\% \\
 \hline
 \multicolumn{2}{|c||}{true unknown rate} & - & 84.9\% \\
 \hline
\multirow{2}*{precision} & known & - & 87.4\% \\
 \cline{2-4}
~ & unknown & - & 91.6\%\\
 \hline
\end{tabular}}
\end{table*}

Apparently, the curves for the primary two kinds of true rate are monotonic, increasing for the known while decreasing for the unknown. The maximum points of these weighted true rate curves for each dataset, are about 0.4, 0.2, and 0.4 respectively. These points exactly correspond to the results shown in Table \ref{table:2}. Besides, the width of the discrimination interval of 2016.10B is only approximately one third of those of 2016.10A and 2016.04C. This implies that the features of 2016.10B are more difficult to learn, and just accounts for its relatively poor performance. 

\subsection{Ablation Study}
In this subsection, we explain the necessity of each of the three loss functions. Relevant experiments are mainly based on 2016.10A. 

Fig. \ref{figure:4} presents the known accuracy in absence of cross entropy loss, center loss and reconstruction loss respectively during training. In general, we found that the best performance in training will be degraded when any one of these three loss functions is excluded. It can be observed that both cross entropy loss and reconstruction loss make a positive impact on the known accuracy, boosting about 3\% to 5\%, while center loss seems slightly weaker. 

Analyzing Table \ref{table:6}, we can easily discern the effect of these three loss functions in the test course, especially the center loss. Results show that the F1 score in absence of cross entropy loss, center loss and reconstruction loss decreases by 1.8\%, 1.7\% and 2.0\% respectively for the known classes. For the unknown classes, the minimum degradation in F1 score is 0.8\% after removing cross entropy loss, while the maximum degradation in F1 score is 16.7\% after removing center loss. Actually, Fig. \ref{figure:3} indicates that the usage of center loss on 2016.10A indeed helps our model to discriminate more distinctly, resulting in a notably broader discrimination interval. Besides, we have also made an attempt at applying L1 loss \cite{7797130} to calculate center loss (Eq. \eqref{ct} in Section \ref{Approach}) and reconstruction loss (Eq. \eqref{ae} in Section \ref{Approach}). Those related results are presented in the last column of Table \ref{table:6}. It is seen that L1 loss can indeed slightly increase the F1 score of known classes by 0.8\%, however, at the cost of a decrease in the F1 score of unknown classes by 6.4\%.


In sum, the three loss functions, though not exactly promoting our \deepmodel~in the same way and in the same fields, are indeed useful.

\subsection{Other Extensions}
We tentatively change several unknown classes on 2016.10A, seeking to excavate more in the feature domain of data. As shown in Table \ref{table:3}, both known precision (KP) and unknown precision (UP) are insensitive to the change of unknown classes, proving that the classification ability of \deepmodel~are consistent and well-preserved for the considered dataset. Nevertheless, obviously, the unknown class CPFSK is always the hardest obstacle in the course of discrimination. Since accuracy of CPFSK is always the lowest as well as some isotopic classes are observed in this case. Especially, when class CPFSK and GFSK simultaneously show up in the unknown roles, the performance loss (on both TKR and TUR) is quite striking. We speculate that the unknown CPFSK and GFSK may share a considerable number of similarities with some known classes, which will unluckily mislead \deepmodel~about the further discrimination task.

To justify \deepmodel's superiority, we compare it with a couple of traditional methods prevailing in the field of outlier detection, as well as two open-set recognition methods, i.e., OpenMax \cite{bendale2016towards} and MDL4OW \cite{liu2020few}. For outlier detection methods, the detected outlier will be regarded as an unknown sample. For OpenMax, an extra dimension is appended to the output vector to indicate the probability of the current sample being unknown. While for MDL4OW, the extreme value theory is adopted to detect the unknown classes by modeling the distribution of loss. The results are presented in Table \ref{table:4}. It is found that our SR2CNN significantly outperforms both outlier detection methods and open-set recognition methods in terms of the true unknown rate. Furthermore, we find that most of the aforementioned methods cannot correctly identify GFSK as unknown. For example, in our experiment, OpenMax wrongly classifies all GFSK samples as known. As for MDL4OW, it identifies a small percentage of GFSK samples at the cost of true known rate. However, it can be found from the experiment results that our SR2CNN can still work very well for this open-set recognition task.

Note that there are no unknown classes identification tasks launched, only discrimination tasks are considered. Hence, here, for a certain unknown class $j$, we compute its unknown rate, instead of accuracy, as $\frac{\#^j_{unknown}}{N_j}$, where $N_j$ denotes the number of samples from unknown class $j$, while $\#^j_{unknown}$ denotes the number of samples from unknown class $j$, which are discriminated as unknown ones. 

\begin{figure}[h]
\includegraphics[width=8.9cm]{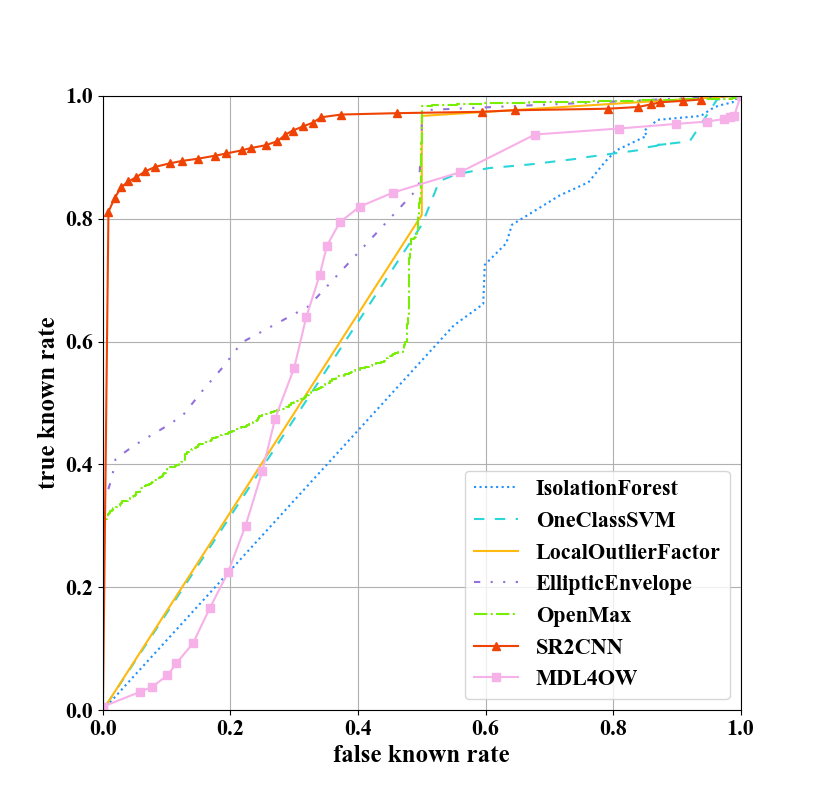}
\caption{ROC curves of \deepmodel, OpenMax and some other outlier detecors on 2016.10A.}
\centering
\label{figure:ROC}
\end{figure}

In addition, relevant ROC curves for the above comparison experiments are depicted in Fig. \ref{figure:ROC}. It is observed that SR2CNN has the largest AUC, indicating its superiority over other methods. Besides, notably, there seems as if a steep `cliff erecting' where False Known Rate approximately equals to 0.5, particularly for EllipticEnvelope, LocalOutlierFactor, and OpenMax. This means that almost half samples of unknown classes are not easy to be correctly discriminated. Correspondingly, according to Table \ref{table:4}, we can speculate that these `hard' samples all come from unknown class GFSK.

\section{Dataset SIGNAL-202002}
We newly synthesize a dataset, denominated as SIGNAL-202002, to hopefully be of great use for further researches in signal recognition field. Basically, the dataset consists of 11 modulation types, which are BPSK, QPSK, 8PSK, 16QAM, 64QAM, PAM4, GFSK, CPFSK, B-FM, AM-DSB and AM-SSB. Each type is composed of 20000 frames. Data is modulated at a rate of 8 samples per symbol, while 128 samples per frame. The channel impairments are modeled by a combination of additive white Gaussian noise, Rayleigh fading, multipath channel and clock offset. We pass each frame of our synthetic signals independently through the above channel model, seeking to emulate the real-world case, which shall consider translation, dilation and impulsive noise, etc. The configuration is set as follows:\par 
~~~~~~20000 samples per modulation type\par
~~~~~~$2\times128$ feature dimension\par
~~~~~~20 different SNRs, even values between [2dB, 40dB]\par
The complete dataset is stored as a python pickle file which is about 450 MBytes in complex 32 bit floating point type. Related code for the generation process is implemented in MatLab and the SIGNAL-202002 dataset is available on the link: \url{https://drive.google.com/file/d/1EDfKRNIk\_txxyAyPCR7BEGs0BvEk3Bof/view}.

We conduct zero-shot learning experiments on our newly-generated dataset and report the results here. As mentioned above, a supervised learning trial is similarly carried out to help us get an overview of the regular performance for each class of SIGNAL-202002. Unfortunately, as Table \ref{table:5} shows, the original two candidates of 2016.10A, AM-SSB and GFSK, both fail to keep on top. Therefore, here, we relocate the unknown roles to another two modulations, CPFSK with the highest accuracy overall, as well as B-FM, which stands out in the three analogy modulation types (B-FM, AM-SSB and AM-DSB).

According to Table \ref{table:5}, an apparent loss on the discrimination ability is observed, as both the TKR and the TUR just slightly pass 80\%. However, our \deepmodel~still maintain its classification ability, as the accuracy for each class remains encouraging compared with the completely-supervised model. A significant fact is that, the known precision (KP) is incredibly high, even exceeding those KPs on 2016.10A by almost 10\%, as shown in Table \ref{table:3}. To account for this, we speculate that the absence of two unknown classes may unintentionally allow \deepmodel~to better focus on the features of the known ones, which consequently, leads to a superior performance of known classification task. 


\section{Conclusion}
In this paper, we have proposed a ZSL framework \deepmodel, which can successfully extract precise semantic features of signals and discriminate both known classes and unknown classes. \deepmodel~can works very well in the situation where we have no sufficient training data for certain class. Moreover, \deepmodel~can generally improve itself in the way of updating semantic center vectors. Extensive experiments demonstrate the effectiveness of \deepmodel. In addition, we provide a new signal dataset SIGNAL-202002 including eight digital and three analog modulation classes for further research. Finally, we would like to point out that, because we often have I/Q signals, a possible direction for future research is using complex neural networks \cite{hirose2013complex} to establish the semantic space.



%

\appendices
\section{Convolution and Deconvolution Operation}
Let $\bm{a},\bm{b} \in \mathbb{R}^n$ denote the vectorized input and output matrices. Then the convolution operation can be expressed as
\begin{equation}
    \bm{b} = \mathbf{M}\bm{a}
\end{equation}
where $\mathbf{M}$ denotes the convolutional matrix, which is sparse. 
With back propagation of convolution, $\frac{\partial Loss}{\partial \bm{b}}$ is obtained, thus
\begin{equation}
    \frac{\partial Loss}{\partial a_j} = \sum_i \frac{\partial Loss}{\partial b_i} \frac{b_i}{a_j} = \sum_i \frac{\partial Loss}{\partial b_i} \mathbf{M}_{i,j} = \mathbf{M}_{*,j}^T \frac{\partial Loss}{\partial \bm{b}}
\end{equation}
where $a_j$ denotes the $j$-th element of $\bm{a}$, $b_i$ denotes the $i$-th element of $\bm{b}$, $\mathbf{M}_{i,j}$ denotes the element in the i-th row and j-th column of $\mathbf{M}$, and $\mathbf{M}_{*,j}$ denotes the $j$-th column of $\mathbf{M}$. Hence,
\begin{equation}
    \frac{\partial Loss}{\partial \bm{a}} = \left[
 \begin{matrix}
  \frac{\partial Loss}{\partial a_1} \\  \frac{\partial Loss}{\partial a_2} \\ \vdots \\  \frac{\partial Loss}{\partial a_n} 
  \end{matrix} \right]= \left[
 \begin{matrix}
  \mathbf{M}_{*,1}^T \frac{\partial Loss}{\partial \bm{b}} \\  \mathbf{M}_{*,2}^T \frac{\partial Loss}{\partial \bm{b}} \\ \vdots \\  \mathbf{M}_{*,n}^T \frac{\partial Loss}{\partial \bm{b}}
  \end{matrix} \right] = \mathbf{M}^T \frac{\partial Loss}{\partial \bm{b}}.
\end{equation}
Similarly, the deconvolution operation can be notated as
\begin{equation}
    \bm{a} = \mathbf{\widetilde{M}}\bm{b}
\end{equation}
where $\mathbf{\widetilde{M}}$ denotes a convolutional matrix that has the same shape as $M^T$, and it needs to be learned. Then the back propagation of convolution can be formulated as follows:
\begin{equation}
    \frac{\partial Loss}{\partial \bm{b}} = \mathbf{\widetilde{M}}^T \frac{\partial Loss}{\partial \bm{a}}.
\end{equation}

For example, the size of the input and output matrices is $4 \times 4$ and $2 \times 2$ as shown in Fig. \ref{3:c}. Then $\bm{a}$ is a 16-dimensional vector and $\bm{b}$ is a 4-dimensional vector. Define convolutional kernel $\mathbf{K}$ as
\begin{equation}
    \mathbf{K} = \left[
 \begin{matrix}
  w_{00} &  w_{01} &  w_{02} \\
  w_{10} &  w_{11} &  w_{12} \\
  w_{20} &  w_{21} &  w_{22}
  \end{matrix}
  \right].
\end{equation}
It is not hard to imagine that $\mathbf{M}$ is a $4 \times 16$ matrix, and it can be represented as follows:
\begin{equation}
\left[
 \begin{matrix}
  w_{00} &  w_{01} &  w_{02} & 0 & \hdots & 0 & 0 & 0 & 0 \\
  0 & w_{00} &  w_{01} &  w_{02} & \hdots & 0 & 0 & 0 & 0 \\
  0 & 0 & 0 & 0 & \hdots & w_{20} &  w_{21} &  w_{22} & 0 \\
  0 & 0 & 0 & 0 & \hdots & 0 & w_{20} &  w_{21} &  w_{22}
  \end{matrix}
  \right].
\end{equation}
Hence, deconvolution is expressed as left-multiplying $\mathbf{\widetilde{M}}$ in forward propagation, and left-multiplying $\mathbf{\widetilde{M}}^T$ in back propagation.





%

\bibliographystyle{IEEEtran}
\bibliography{paper}

\begin{IEEEbiography}[{\includegraphics[width=1in,height=1.25in,clip,keepaspectratio]{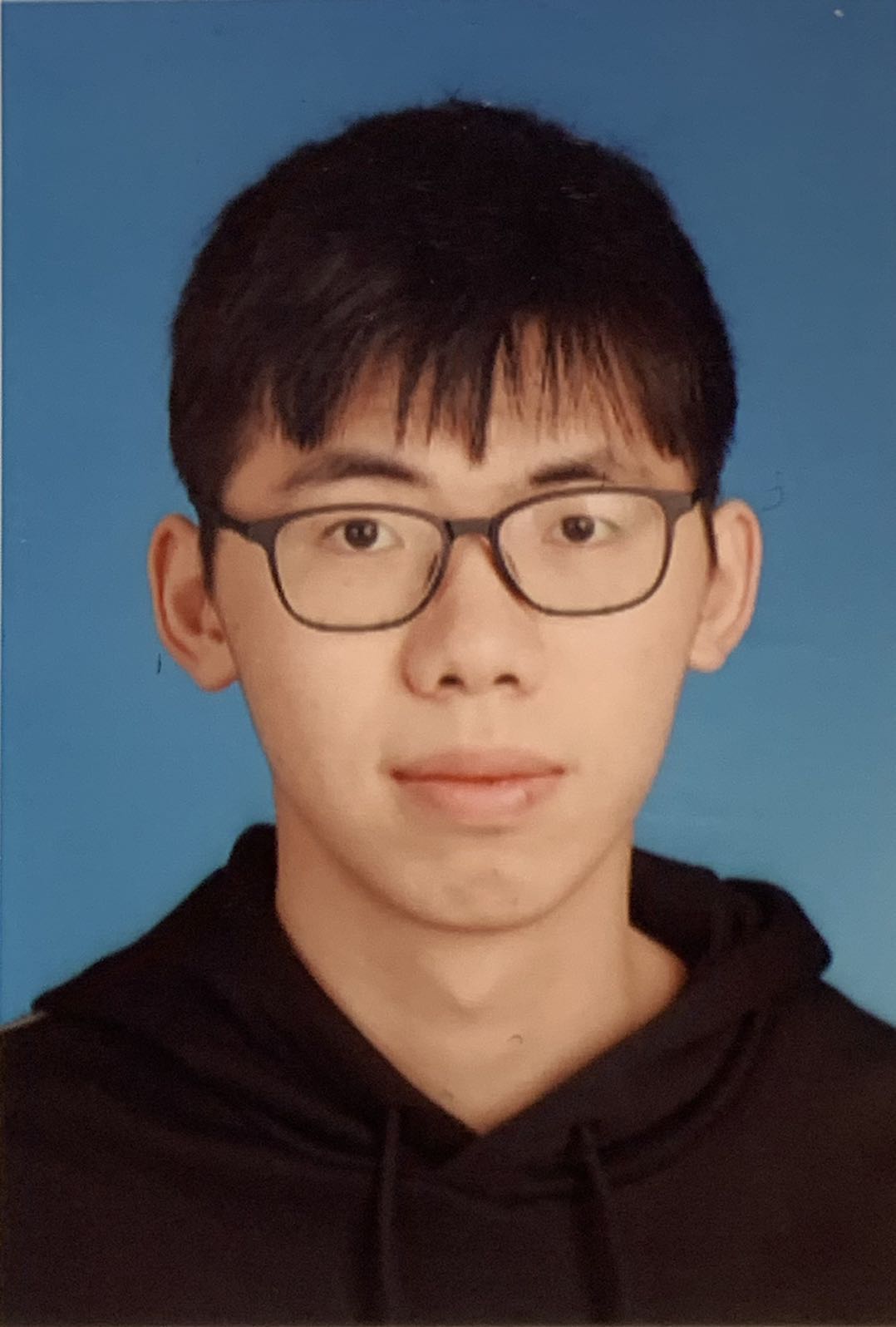}}]{Yihong Dong}
received his B.S. degree in Computer Science from Shanghai University, China, in 2019. He is currently a graduate student with the School of Software Engineering, Tongji University. His research interests include machine learning with the applications in signal processing.
\end{IEEEbiography}

\begin{IEEEbiography}[{\includegraphics[width=1in,height=1.25in,clip,keepaspectratio]{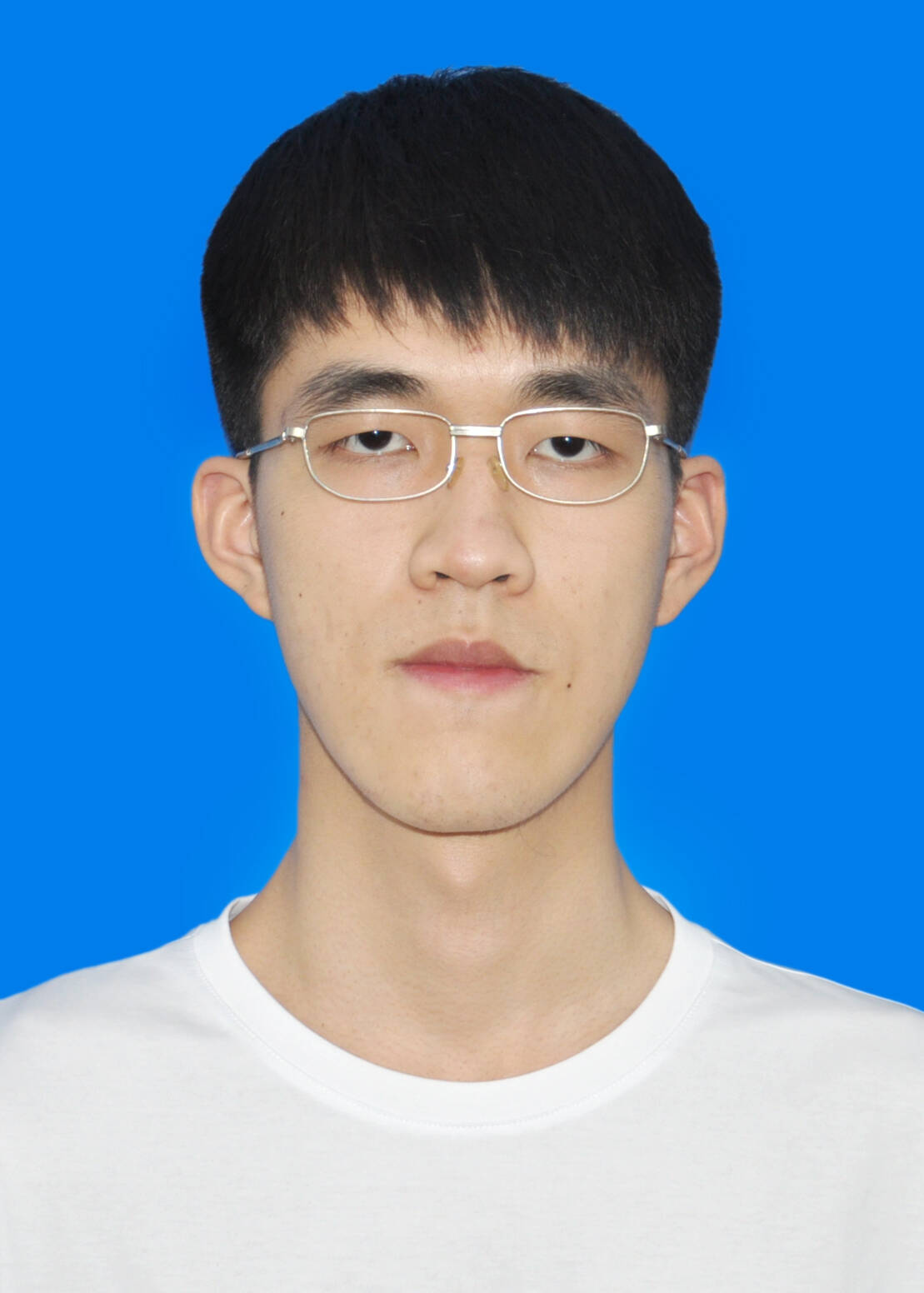}}]{Xiaohan jiang}
is currently an undergraduate with the School of Software Engineering, Tongji University. His research interests include deep learning and computer vision.
\end{IEEEbiography}

\begin{IEEEbiography}[{\includegraphics[width=1in,height=1.25in,clip,keepaspectratio]{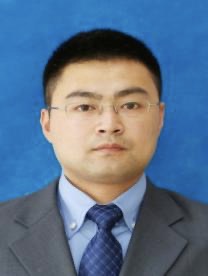}}]{Huaji Zhou}
was born in Jinhua, Zhejiang, China, in 1988. He received the B.S. degree in Navigation, Guidance, and Control Technology and the M.S. degree in Pattern Recognition and Intelligent System from Xidian University, Xi’an, China, in 2010 and 2013, respectively. 
He is currently a Research Assistant with Science and Technology on Communication Information Security Control Laboratory, Jiaxing, China. And he is pursuing Ph.D. degree in Electronics and Information at Xidian University, Xi'an, China, His current research interests include machine learning and electromagnetic signal processing.
\end{IEEEbiography}

\begin{IEEEbiography}[{\includegraphics[width=1in,height=1.25in,clip,keepaspectratio]{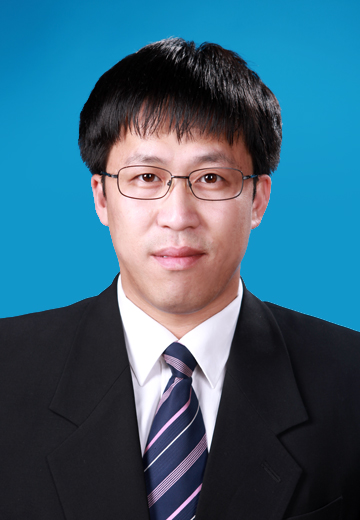}}]{Yun Lin}
received the B.S. degree from Dalian Maritime University, Dalian, China, in 2003, the M.S. degree from the Harbin Institute of Technology, Harbin, China, in 2005, and the Ph.D. degree from Harbin Engineering University, Harbin, China, in 2010. He was a research scholar with Wright State University, USA, from 2014 to 2015. Now, he is currently a full professor in the College of Information and Communication Engineering, Harbin Engineering University, China. His current research interests include machine learning and data analytics over wireless networks, signal processing and analysis, cognitive radio and software defined radio, artificial intelligence and pattern recognition. He had published more than 150 international peer-reviewed journal/conference papers, such as the IEEE IoT, TII, TVT, TCCN, TR, Access, INFOCOM, GLOBECOM, ICC, VTC, ICNC. He had four high-cited papers and several best conference papers. He is serving as an editor for the IEEE TRANSACTIONS ON RELIABILITY, KSII Transactions on Internet and Information Systems, and International Journal of Performability Engineering. In addition, he served as General Chair of ADHIP 2020, TPC Chair of MOBIMEDIA 2020, ICEICT 2019 and ADHIP 2017, and TPC member of GLOBECOM, ICC, ICNC and VTC. He had successfully organized several international workshops and symposia with top‐ranked IEEE conferences, including INFOCOM, GLOBECOM, DSP, ICNC, among others.
\end{IEEEbiography}

\begin{IEEEbiography}[{\includegraphics[width=1in,height=1.25in,clip,keepaspectratio]{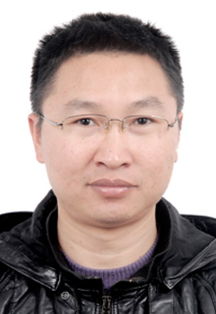}}]{Qingjiang Shi}
received his Ph.D. degree in electronic engineering from Shanghai Jiao Tong University, Shanghai, China, in 2011. From September 2009 to September 2010, he visited Prof. Z.-Q. (Tom) Luo's research group at the University of Minnesota, Twin Cities. In 2011, he worked as a Research Scientist at Bell Labs China. From 2012, He was with the School of Information and Science Technology at Zhejiang Sci-Tech University. From Feb. 2016 to Mar. 2017, he worked as a research fellow at Iowa State University, USA. From Mar. 2018, he is currently a full professor with the School of Software Engineering at Tongji University. He is also with the Shenzhen Research Institute of Big Data. His interests lie in algorithm design and analysis with applications in machine learning, signal processing and wireless networks. So far he has published more than 60 IEEE journals and filed about 30 national patents. Dr. Shi was an Associate Editor for the IEEE TRANSACTIONS ON SIGNAL PROCESSING. He was awarded Golden Medal at the 46th International Exhibition of Inventions of Geneva in 2018, and also was the recipient of the First Prize of Science and Technology Award from China Institute of Communications in 2018, the National Excellent Doctoral Dissertation Nomination Award in 2013, the Shanghai Excellent Doctorial Dissertation Award in 2012, and the Best Paper Award from the IEEE PIMRC'09 conference.

\end{IEEEbiography}







\end{document}